\documentclass[twocolumn]{autart}

\usepackage{graphicx}
\usepackage{caption}
\usepackage{amssymb}

\usepackage{multicol}
\usepackage{amsmath,amssymb}
\usepackage{amsfonts}
\usepackage{textcomp}
\usepackage{psfrag}
\usepackage{multimedia}
\usepackage{fancybox}

\newcommand{\vect}[1]{\ensuremath{\boldsymbol{\mathrm{#1}}}}

\newtheorem{theorem}{Theorem}

\newtheorem{Lemma}{Lemma}
\newtheorem{Definition}{Definition}
\newtheorem{Remark}{Remark}

\newtheorem{Assumption}{Assumption}

\usepackage[usenames,dvipsnames]{xcolor}
\definecolor{wheat}{rgb}{0.96,0.87,0.70}
\definecolor{mario}{rgb}{0.8,0.8,1}
\definecolor{seb}{rgb}{0.8,1,0.8}

\newcommand {\matr}[2]{\left[\begin{array}{#1}#2\end{array}\right]}

\newcommand{\norm}[1]{\left \|\begin{matrix}#1\end{matrix}\right \|}
\usepackage{comment}

\newcounter{lastnote}

\usepackage[algo2e,norelsize,ruled,vlined,linesnumbered]{algorithm2e}

\begin{document}

\begin{frontmatter}

\title{Learning for MPC with Stability \& Safety Guarantees\thanksref{footnoteinfo} } 

\thanks[footnoteinfo]{This paper was not presented at any IFAC 
meeting. Corresponding author M. Zanon. This paper was partially supported by the Italian Ministry of University and Research under the PRIN'17 project "Data-driven learning of constrained control systems" , contract no. 2017J89ARP; and by the Norwegian Research Council project ``Safe Reinforcement-Learning using MPC" (SARLEM).}

\author[Seb]{Sebastien Gros}\ead{sebastien.gros@ntnu.no},
\author[Mario]{Mario Zanon}\ead{mario.zanon@imtlucca.it}

\address[Seb]{Dept. of Cybernetics, Faculty of Information Technology, NTNU, Norway}
\address[Mario]{IMT School for Advanced Studies Lucca, Piazza San Francesco 19, 55100, Lucca, Italy}


\begin{abstract}
The combination of learning methods with Model Predictive Control (MPC) has attracted a significant amount of attention in the recent literature. The hope of this combination is to reduce the reliance of MPC schemes on accurate models, and to tap into the fast developing machine learning and reinforcement learning tools to exploit the growing amount of data available for many systems. In particular, the combination of reinforcement learning and MPC has been proposed as a viable and theoretically justified approach to introduce explainable, safe and stable policies in reinforcement learning. However, a formal theory detailing how the safety and stability of an MPC-based policy can be maintained through the parameter updates delivered by the learning tools is still lacking. This paper addresses this gap. The theory is developed for the generic Robust MPC case, and applied in simulation in the robust tube-based linear MPC case, where the theory is fairly easy to deploy in practice. The paper focuses on Reinforcement Learning as a learning tool, but it applies to any learning method that updates the MPC parameters online.
\end{abstract}

\end{frontmatter}

\section{Introduction} 

Model Predictive Control (MPC) is a very successful tool for generating policies that minimize a certain cost under some state and input constraints. MPC uses model-based predictions of the future system trajectories to produce a control input profile over a future time window that satisfies the constraints while minimizing the cost. Closed-loop control policies are then obtained by updating that control input profile at every time instant, in a receding-horizon fashion, based on the latest state of the system and information on its environment. MPC heavily relies on a model of the system at hand to perform well. However, accurate models are expensive to develop, and can be too complex to use in the context of MPC. As a result, while MPC can deliver a reasonable approximation of the optimal policy, it is usually suboptimal. 

In the recent literature, various learning techniques, often borrowed from the field of machine learning, have been investigated to address this problem. A number of them focus on using machine learning to improve the fitting of the MPC model to the data, see, e.g.,~\cite{Hewing2020} and references therein.  Other approaches argue for adjusting the MPC model, cost and constraints in view of maximizing directly the closed-loop performance. The authors of \cite{Gros2020} provide strong theoretical justifications for this approach, and propose to use Reinforcement Learning (RL) techniques to perform that adjustment in practice. The use of learning techniques within control has been proposed in, 
e.g.,~\cite{Aswani2013,Berkenkamp2017,Koller2018,Lewis2009,Lewis2012,Murray2018,Ostafew2016,Amos2018,Gros2020,Zanon2021,Zanon2019,Masti2022,Zhu2022,Abdufattokhov2021,Wabersich2019,Soloperto2023}.

A prime motivation for using MPC-based policies is the possibility to enforce constraints on the system trajectories. This feature can be leveraged to ensure the safety of the system at hand by defining constraints that limit its evolution to a safe set, and including these constraints in the MPC formulation. When using inaccurate models, or when the real system is stochastic, safety can be maintained via robust MPC techniques, where the uncertainties are taken into account in the MPC formulation. Robust MPC delivers policies that are safe by construction, using a worst-case approach, and ensures that the real system trajectories satisfy the constraints at all time.

The combination of learning with robust MPC has been investigated in~\cite{Zanon2021}. This combination arguably offers the most direct pathway to optimize the closed-loop performance of an MPC-based policy while maintaining its safety. The authors of~\cite{Zanon2021} additionally argue that robust MPC offers a direct pathway towards safe reinforcement learning, providing strong certificates of safety. Indeed, when using robust MPC, the policies produced via learning can be made safe and stable by construction by imposing certain constraints on the parameter updates suggested by the learning algorithm~\cite{Zanon2021}. In contrast, in classic reinforcement learning, e.g., based on Deep Neural Networks (DNN), enforcing the safety of the resulting policy is typically done via extensive in silico validations using Monte Carlo techniques. In that context, formal certificates of strong safety require in principle an infinite amount of simulations and are therefore difficult to establish in practice.

While \cite{Zanon2021} details how to learn policies that are safe and stable by construction, an important gap remains to be addressed. Performing learning on a control policy requires that regular updates of the policy parameters are implemented. Some learning methods implement the updates at every sampling time of the policy, while other methods implement the updates less frequently. If the parameter updates are to be implemented while the system is being operated, then implementing safe and stable policies at every parameter update does not necessarily yield an overall safe and stable closed-loop system. Indeed, safety with parameter updates can only be guaranteed if the update takes place when the system state is within specific sets. Stability is also not guaranteed when the parameters are frequently updated, even if each policy implemented is stable. Ensuring safety and stability through the parameter updates then requires additional conditions. This paper addresses that issue by detailing how to maintain safety and stability through the parameter updates, and therefore provides a complete theoretical framework to deploy safe and stable learning for MPC. The paper focuses on learning techniques based on reinforcement learning, but applies equally to all learning techniques that propose directions in the MPC parameters space along which the MPC parameters ought to be updated.

The paper is organized as follows. Section~\ref{sec:Background} provides background material on MDPs. Section~\ref{sec:SafeRL} proposes a definition of safe policies and Section~\ref{sec:StableRLMPC:generic} provides background material on robust MPC. Section~\ref{sec:RecFeas} presents conditions for building a safe combination of RL and robust MPC, where the parameter updates provided by RL do not jeopardize the safety of the robust MPC scheme. Section~\ref{sec:MPCStability} presents conditions for RL to update the MPC parameters while maintaining the system stability, discussed in an augmented parameter-state space. Section~\ref{sec:simulations} illustrates the theoretical results by means of two simple examples and Section~\ref{sec:conclusions} concludes the paper.

\section{Background} \label{sec:Background}

We consider real systems that can be described as discrete-time Markov chains with continuous state and action spaces. We will label the underlying conditional transition probability density over the states $\vect s$ and actions $\vect a$ as:
\begin{align}\label{eq:Transition}
\varphi \left[\,\vect s_{i+1} \,|\, \vect s_i,\vect a_i\right]\,:\, \mathbb R^n \times  \mathbb R^n \times  \mathbb R^m\,\rightarrow \mathbb R_+,
\end{align}
where $n$ and $m$ are the state and input space sizes, respectively. Throughout the paper, index $i$ will refer to the physical time of the system. We will assume that \eqref{eq:Transition} is only inaccurately known. We will assume in the following that a stage cost 
\begin{align}
L\left(\vect s_i,\vect a_i\right)\,:\, \mathbb R^n \times  \mathbb R^m\,\rightarrow \mathbb R
\end{align}
is provided, and that our goal is to find the parameters $\vect\theta$ of a policy 
\begin{align}
\vect \pi_{\vect\theta}\,:\, \mathbb R^n \rightarrow \mathbb R^m
\end{align}
delivering the actions, i.e., $\vect a_i = \vect \pi_{\vect\theta}\left(\vect s_i\right)$, so as to minimize the expected discounted cost:
\begin{align}
\label{eq:JIntro}
J\left(\vect \pi_{\vect\theta} \right) = \mathbb E\left[\left.\, \sum_{i=0}^\infty\,\gamma^i L\left(\vect s_i,\vect a_i\right)\,\right|\, \vect a_i = \vect\pi_{\vect\theta}\left(\vect s_i\right)\right],
\end{align}
where $\mathbb E[\cdot]$ is the expected value operator applying to the real trajectories yielded by \eqref{eq:Transition} in closed-loop with policy $\vect \pi_{\vect\theta}$, and $\gamma\in (0,1]$ a discount factor. We will consider that the actions delivered by the policy are possibly restricted to a subset of $\mathbb R^m$, i.e., the minimization of $J$ is subject to
\begin{align}
\label{eq:Pol:intro}
\vect \pi_{\vect\theta} \left(\vect s_i\right)\in \mathbb P,\quad \forall\,\vect s_i.
\end{align}
Finding the parameters $\vect \theta$ that (locally) minimize the closed-loop cost $J\left(\vect \pi_{\vect\theta} \right)$, and therefore maximize the closed-loop performance of the policy is arguably one of the main goals of any learning algorithms focusing on improving a policy. E.g., many methods in Reinforcement Learning (RL) deal with the evaluation of the policy gradient $\nabla_{\vect\theta}J\left(\vect \pi_{\vect\theta} \right)$, which is used to update the policy parameters $\vect\theta$ such that $J\left(\vect \pi_{\vect\theta} \right)$ is sequentially decreased. In this paper, we will focus for simplicity on parameter updates that reduce $J\left(\vect \pi_{\vect\theta} \right)$ directly, albeit a number of learning techniques compute updates that are not directly based on  $J\left(\vect \pi_{\vect\theta} \right)$. The theory presented here applies to all learning methods for MPC with minor modifications.

We consider in this paper that we seek policies that keep the system safe in the sense of respecting some state constraints expressed as:
\begin{align}
\label{eq:SafeConst}
\vect s_i \in\mathcal X,
\end{align}
for any time $i= 0,\ldots,\infty$, where $\vect s_{0,\ldots, \infty}$ denotes the closed-loop trajectories with policy $\vect \pi_{\vect\theta}$. Note that for the sake of simplicity, we will not treat mixed state-input constraints here, even though the proposed results arguably readily extend to that case. Throughout the paper, we will assume that the real state transition \eqref{eq:Transition} is imperfectly or only coarsely known, and difficult to capture via simple mathematical models.

\section{Safe Policies} \label{sec:SafeRL}
Ideally, a policy is safe if \eqref{eq:SafeConst} holds at all time $i$ with probability one. Unfortunately, guaranteeing this unitary probability is impossible without a perfect knowledge of the system dynamics \eqref{eq:Transition}. In that context, a more realistic notion of safety can be defined in the context of Bayesian inference, where safety is regarded as probabilistic, conditioned on our knowledge of the system (prior and actual data), labeled $\mathcal D$. Such data can, e.g., be the set of all state transitions $\vect s, \vect a, \vect s_+ $  observed so far, but also include some prior knowledge of the system. The strictest notion of safety hence becomes a probabilistic counterpart of~\eqref{eq:SafeConst}, 
which can be defined in terms of constraint satisfaction as: 
\begin{align}
\label{eq:sigmasafety0}
\sigma := \mathbb P\left[ \,\vect s_i \in\mathcal X\quad \forall i\,|\, \mathcal D\,\right].
\end{align}
Probability \eqref{eq:sigmasafety0} is epistemological, and to be understood in the context of Bayesian hypothesis testing. It underlines that safety can only be a belief conditioned on our current knowledge of the system at hand. In this paper, we adopt that operational notion of knowledge-based safety and label a policy satisfying \eqref{eq:sigmasafety0} as $\sigma$-safe. Assessing probability \eqref{eq:sigmasafety0} in practice can be difficult, but it can arguably be done in several ways:
\begin{enumerate}
\item[1.] Direct data-based: data $\mathcal D$ are collected on the real system, and used to infer an estimation of $\sigma${, without using a model of~\eqref{eq:Transition}}. The obvious difficulty here lies in the need for collecting an extremely large data set if a policy with $\sigma$ close to one is to be designed. Real data are costly, especially for safety-critical systems, and designing policies that achieve a high $\sigma$ can be unrealistic in that context.
\item[2.] Direct model-based: a ``pessimistic" simulation model of the real system is constructed (see Equations~\eqref{eq:StateTraj}-\eqref{eq:IncludedSupport} below) from $\mathcal D$, and $\sigma$ is estimated in silico via Monte Carlo methods. If the model is pessimistic, the in-silico estimation of $\sigma$ converges (as more in-silico data are generated) to a lower bound for the true $\sigma$.
\item[3.] Indirect model-based: similarly to 2, a ``pessimistic" control model of the real system is constructed, and used to build policies that are safe by construction for that model. Monte Carlo sampling is here replaced by an explicit or implicit propagation of the uncertainty set, based on an overestimation of the support of \eqref{eq:Transition}. The resulting $\sigma$ is by construction a lower bound for \eqref{eq:sigmasafety0}.
\end{enumerate}

Examples of approach 1. can be found in, e.g.,~\cite{Gaskett2003,Heger1994}; examples of approach 2. can be found in~\cite{Papaioannou2015,Zuev2021}; and examples of approach 3. include, e.g.,~\cite{Chisci2001,Mayne2005,Villanueva2017}. 
While approach 2. is often used in the context of safe reinforcement learning (providing only weak guarantees that hold upon convergence) and to certify control policies in practice, it is difficult to provide strong safety guarantees in that context. In this paper, we will consider the third approach, based on robust MPC techniques. Our intentions are two-fold. First, we present a theory that specifies formally how learning can be deployed in the MPC context without jeopardizing the closed-loop stability and safety of the resulting policy throughout the learning process. Second, we propose this framework as one viable approach to safe reinforcement learning if strong safety and stability guarantees are expected to be satisfied throughout the learning process.

The models of the real system used in approaches~2. and~3. above are often built as structured models that can, e.g., take the form:
\begin{align}
\label{eq:StateTraj}
\vect s_{i+1} = \vect F_{\vect\vartheta}\left(\vect s_i, \vect a_i, \vect w_i\right),
\end{align}
where $\vect F_{\vect\vartheta}\,:\, \mathbb R^n \times  \mathbb R^m \times  \mathbb R^d\,\rightarrow \mathbb R^n $, and $\vect\vartheta$ is the set of model parameters. Moreover, variable
\begin{align}
\label{eq:DistSet}
\vect w_i\in\mathbb W_{\vect\vartheta}\subset\mathbb R^d,
\end{align}
is an external, typically stochastic disturbance contained in a compact set $\mathbb W_{\vect\vartheta}$, modeling the stochasticity of the real system. The notion of ``pessimistic" model used above then requires that for any state-action pair $\vect s,\vect a$, the support of the real system dynamics density \eqref{eq:Transition} is (almost entirely) included in the set: 
\begin{align}
\label{eq:SupportingSet}
&\mathbb D_{\vect\vartheta}\left(\vect s,\vect a\right) = \left\{\, \vect F_{\vect\vartheta}\left(\vect s, \vect a, \vect w\right) \,|\, \forall\,\vect w \in\mathbb W_{\vect\vartheta}\right\},
\end{align}
i.e.,
\begin{align}
\label{eq:IncludedSupport}
&\int_{\mathbb D_{\vect\vartheta}\left(\vect s,\vect a\right)} \varphi \left[\,\vect s_{+} \,|\, \vect s,\vect a\right]\mathrm d\vect s_+ = 1,\quad \forall\, \vect s,\vect a.
\end{align}
Condition \eqref{eq:IncludedSupport} provides a formal definition of a ``pessimistic" model, i.e., a model that includes the support of the real state transition \eqref{eq:Transition}. Note that this condition does not necessarily need to be conservative. Indeed \eqref{eq:IncludedSupport} can in principle hold tightly, with $\mathbb D_{\vect\vartheta}$ covering the support of $\varphi$ but not more. Condition \eqref{eq:IncludedSupport} further entails that a policy guaranteeing that \eqref{eq:SafeConst} is satisfied for all the possible trajectories resulting from \eqref{eq:StateTraj}-\eqref{eq:DistSet} is safe by construction. Robust MPC techniques can be used to build such policies.

Similarly to \eqref{eq:sigmasafety0}, the validity of  \eqref{eq:IncludedSupport} is limited to our knowledge of the system supported by all data and prior knowledge available about it, i.e., $\mathcal D$. Condition \eqref{eq:IncludedSupport} then ought to be regarded as probabilistic as well, in which case it becomes:
\begin{align}
\label{eq:sigmasafety}
\tilde\sigma := \mathbb P\left[ \,\eqref{eq:IncludedSupport}\,|\, \mathcal D\,\right].
\end{align}
One can easily verify that if a policy \eqref{eq:Pol:intro} ensures that the closed-loop trajectories of the model \eqref{eq:StateTraj}-\eqref{eq:IncludedSupport} respect the state constraints \eqref{eq:SafeConst} at all time, then the probability that the policy is safe for the real system is at least $\tilde\sigma$, and $\sigma \geq \tilde\sigma$ holds. In practice, a minimum requirement for $\tilde\sigma > 0$ to hold is to ensure that:
\begin{align}
\label{eq:SafeInclusion}
\vect s_+ \in \mathbb D_{\vect\vartheta}\left(\vect s,\vect a\right)
\end{align}
holds for all observed state transition triplets $\left(\vect s, \vect a, \vect s_+\right) \in\mathcal D $. Condition \eqref{eq:SafeInclusion} then defines a set
\begin{align*}
	\vect\Theta_\mathcal D :=\left \{\, \vect{\vartheta} \ | \ \vect{s}_+\in \mathbb D_{\vect\vartheta}\left(\vect s,\vect a\right), \ \forall \, \left(\vect s, \vect a, \vect s_+\right) \in\mathcal D \,\right \}
\end{align*}
where the model parameters $\vect\vartheta$ should be restricted, which is typically tackled via set-membership system identification methods \cite{Bertsekas1971a}. In this paper we will consider a set $\vect\Theta_\mathcal D $ possibly formed from \eqref{eq:SafeInclusion}, and possibly further restricted by prior or structural knowledge of the system. For a complete discussion on the definition of $\vect\Theta_\mathcal D $ based on~\eqref{eq:SafeInclusion} and its deployment within safe RL, we refer to~\cite{Zanon2021}.

\section{Robust MPC as a safe \& stable policy \label{sec:StableRLMPC:generic}}
We will consider the use of robust MPC as a means to generate safe policies $\vect \pi_{\vect\theta} \left(\vect s_i\right)$ 
 for system \eqref{eq:Transition}, subject to the associated performance index \eqref{eq:JIntro} and safety restriction \eqref{eq:SafeConst}, with the limitation \eqref{eq:sigmasafety}. Note that these policies are parametrized by parameter $\vect\theta$, which defines the robust MPC scheme as we will discuss in the following. A strong point of robust MPC is that safety in the sense of Section \ref{sec:SafeRL} and stability can be enforced by construction. A non-trivial remaining question is then how to retain safety and stability through the learning process, where the parameters of the robust MPC scheme are regularly updated.

A generic robust MPC scheme is based on predicting the system evolution at future times based on actions given by a sequence of future, parametrized policies:
\begin{align}
\label{eq:Policy}
\vect a_k = \vect \eta^k_{\vect\theta}\left(\vect v, \vect s_k\right),
\end{align}
where the indices $k$ refer to future time instances, occurring at the corresponding physical times $i+k$ and $\vect v$ is a set of variables used to shape the policy sequence for the specific current state of the system $\vect s_i$. In this paper, we consider robust MPC schemes of the form:
\begin{subequations}
	\label{eq:RNMPC}
	\begin{align}
	\hspace{-3pt}\hat V_{\vect\theta}(\vect s_i) =\min_{\vect v}&\quad \varphi_{\vect\theta}(\vect v,\vect s_i)\label{eq:RCost}\\
	\mathrm{s.t.}&\quad \mathbb X_{k+1} = \vect F_{\vect\vartheta}\left(\mathbb X_k, \vect \eta^k_{\vect\theta}\left(\vect v, \mathbb X_k\right),  \mathbb W_{\vect\theta}\right), \hspace{-10em}&&\hspace{5em} \label{eq:SetTraj}\\
	&\quad \mathbb X_0 = \vect s_i, && \mathbb X_{N}^{\vect{\theta}}\subseteq \mathcal X_{\vect\theta}^\mathrm f, \label{eq:RMPCFinalSet}\\
	&\quad \vect \eta^k_{\vect\theta}\left(\vect v, \mathbb X_k\right)\subseteq\mathbb U,&& \mathbb X_{k}^{\vect{\theta}}\subseteq \mathcal X,\\
	&\quad \forall\, k = 0,\ldots,N-1,
	\end{align}
\end{subequations}
where the state propagation \eqref{eq:SetTraj}, called \emph{tube}, is the extension of  \eqref{eq:StateTraj} to set propagation, i.e., we read  \eqref{eq:SetTraj} as:
\begin{align}
\mathbb X_{k+1}^{\vect{\theta}} = \left\{\, \vect F_{\vect\vartheta}\left(\vect s, \vect \eta^k_{\vect\theta}\left(\vect v, \vect s\right), \vect w\right)\,|\,\forall \, \vect s\in\mathbb X_k^{\vect{\theta}},\,\vect w \in \mathbb W_{\vect\theta} \right\}.
\end{align}
The robust MPC scheme \eqref{eq:RNMPC} defines a policy $\vect\pi_{\vect\theta}\left(\vect s\right)$ given by the first control input of the policy sequence adopted in the robust MPC scheme, i.e.:
\begin{align}
\label{eq:RMPC:Policy}
\vect\pi_{\vect\theta}\left(\vect s_i\right) = \vect \eta^0_{\vect\theta}\left(\vect v^\star, \vect s_i\right),
\end{align}
where $\vect v^\star$ is the solution of \eqref{eq:RNMPC}.
The robust MPC model used in \eqref{eq:SetTraj} is then an intrinsic and central component of the robust MPC scheme formulation. We will then consider that the model parameters $\vect\vartheta$ are part of the policy parameters $\vect\theta$, as they can be used to shape the policy $ \vect \pi_{\vect\theta} $ delivered by the robust MPC scheme.

Set $\mathbb U$ in \eqref{eq:RNMPC} represents the possible limitations on the feasible actions (e.g., actuator limitations), and the terminal set $\mathcal X_{\vect\theta}^\mathrm f$ must be constructed such that \eqref{eq:RNMPC} being feasible entails that the state constraint $\mathbb X_{k}\subseteq \mathcal X$ can be enforced at all future times. This is typically achieved by resorting to a terminal control law which makes $\mathcal X_{\vect\theta}^\mathrm f$ forward invariant. 
The cost function \eqref{eq:RCost} is left unspecified here, as it can take different forms such as, e.g., a worst-case cost (as in min-max robust MPC); a nominal cost (as in tube MPC); an expected cost, or more elaborate risk-adverse costs.
Function $\hat V_{\vect\theta}(\vect s) $ receives an infinite value for the states $\vect s$ for which problem \eqref{eq:RNMPC} is infeasible. For the sake of simplicity, we do not consider mixed state-input constraints here, although the proposed theory readily applies to that case.

We ought to stress that any Robust MPC formulation can be considered in our framework, e.g., robust MPC~\cite{Mayne2005,Chisci2001,Zanon2021a}, possibly also combined with non-standard formulations such as~\cite{Limon2008}. Note that while nonlinear robust MPC is in general hard to formulate and solve, recent research has provided methods that can be applied in practice, see, e.g.,~\cite{Mayne2011,Villanueva2017,Koehler2021}. 
Furthermore, it is arguably useful here to stress that the choice of using robust MPC to carry safe and stable policies, while not necessarily straightforward to implement, is nonetheless not restrictive. Indeed, the existence of a safe policy for the model \eqref{eq:StateTraj}-\eqref{eq:DistSet} entails the existence of a robust MPC scheme~\eqref{eq:RNMPC} delivering a safe policy for the system model~\eqref{eq:StateTraj}-\eqref{eq:DistSet}, and therefore a $\sigma$-safe policy for the real system~\eqref{eq:Transition}. While these observations are well understood in the MPC community, we provide hereafter a Lemma that highlights their importance in a learning context.
\begin{Lemma} 
	\label{lem:safe_stable_policies}
	Assume that there exists a feasible policy $\vect\pi$ such that the trajectories of the model \eqref{eq:StateTraj}-\eqref{eq:DistSet}: (i)  are in $\mathcal X$ with probability 1 for a set of initial conditions $\vect s_0 \in \mathcal X^0$; and (ii) are asymptotically stabilized to some set $\mathbb{L}$ with an associated Lyapunov function $\mathcal{V}_{\vect{\pi}}(\vect{s})$. Then there exists a robust MPC scheme of the form \eqref{eq:RNMPC} that is $\sigma$-safe for the real system dynamics \eqref{eq:Transition}, with probability at least $\sigma$ of being stable for the real system dynamics \eqref{eq:Transition}.
\end{Lemma}
\begin{Remark}
Before delivering the proof of this simple Lemma, it ought to be stressed here that its purpose is not to specify how the robust MPC scheme should be built, but rather to dismiss potential concerns that robust MPC is a limited tool for producing safe and stable policies. Indeed, it shows that if a safe and stable policy exists, then a robust MPC scheme that produces a safe and stable policy also exists. 
 \end{Remark}
\begin{pf} 
	For the first claim it is sufficient to prove that is it possible to setup a recursively feasible robust MPC scheme such that the model dynamics satisfy \eqref{eq:SafeConst}. 
	Let us select $ \vect \eta^k_{\vect\theta}$ as
\begin{align}
 \vect \eta^k_{\vect\theta}\left(\vect v, \vect s \right) = \vect\pi\left(\vect s \right) +  \vect \rho^k_{\vect\theta}\left(\vect v, \vect s \right), \label{eq:RobustMPCPolicyChoice}
\end{align}
where $\vect \rho^k_{\vect\theta}$ is an arbitrary policy such that $ \vect \rho^k_{\vect\theta}\left(0, \vect s \right) = 0$ for all $\vect s$, and select 
\begin{align}
\mathcal X_{\vect\theta}^\mathrm f = \mathcal X^0.
\end{align}
Then constraints \eqref{eq:SetTraj} and \eqref{eq:RMPCFinalSet} ensure that the robust MPC policy maintains the trajectories of the model \eqref{eq:StateTraj}-\eqref{eq:DistSet} in the set $\mathcal X$ and that the robust MPC scheme is recursively feasible. Moreover, the choice of policy \eqref{eq:RobustMPCPolicyChoice} ensures that the constraints are feasible because the trivial choice $\vect v=0$ is feasible.

Concerning the second claim, we exploit the fact that the policy is stabilizing and there exists a Lyapunov function $\mathcal{V}_{\vect{\pi}}(\vect{s})$ associated with set $\mathbb{L}$. By selecting the cost as
\begin{align*}
	 \varphi_{\vect\theta}(\vect v,\vect s) = \mathcal{V}_{\vect{\pi}}(\vect{s}) + \vect v^\top \vect v,
\end{align*}
we obtain that $\vect \eta^k_{\vect\theta}\left(0, \vect s \right)$ is optimal and the MPC value function is a Lyapunov function, i.e., it satisfies $\hat V_{\vect{\theta}}(\vect{s})=\mathcal{V}_{\vect{\pi}}(\vect{s})$. Since policy $\vect{\pi}$ is safe and stabilizing by assumption, the corresponding robust MPC scheme is safe and stable for the model dynamics \eqref{eq:StateTraj}-\eqref{eq:DistSet}. Then, using \eqref{eq:sigmasafety}, we conclude that the resulting policy is $\sigma$-safe for the real system dynamics \eqref{eq:Transition} and stabilizing with probability $\sigma$. 
$\hfill\square$
\end{pf}
Lemma \ref{lem:safe_stable_policies} is further discussed in Remark \ref{rem:lemma1} below.

In practice, for simplicity, the parametrized policy \eqref{eq:Policy} used in \eqref{eq:RNMPC} is often selected as a nominal state-input reference sequence $\bar{\vect u}_{0,\ldots,N-1},\, \bar{\vect x}_{0,\ldots,N-1}$ together with an additional linear state feedback, i.e.,
\begin{align}
	\label{eq:mpc_tube_policy}
	\vect \eta^k_{\vect\theta}\left(\vect v, \vect s\right) &= \bar{\vect u}_k - K_{\vect\theta}\left(\vect s - \bar{\vect x}_k\right), \\
\label{eq:mpc_nominal_trajectory}
\vect v &= \left\{\bar{\vect u}_{0,\ldots,N-1}, \bar{\vect x}_{0,\ldots,N-1}\right\}.
\end{align}
In that specific case, the policy \eqref{eq:RMPC:Policy} defined by the robust MPC scheme becomes 
\begin{align}
\vect\pi_{\vect\theta}\left(\vect s_i\right) = \vect \eta^0_{\vect\theta}\left(\vect v^\star, \vect s_i\right) = \vect{\bar u}_0^\star.
\end{align}
We define next some important sets in the space of parameters $\vect \theta$. 
\begin{Definition} \label{Def:RecFeas} Let us define the set $\vect \Theta_\mathrm F$ of parameters $\vect\theta$ such that the robust MPC scheme \eqref{eq:RNMPC} is recursively feasible for the dynamics \eqref{eq:StateTraj}-\eqref{eq:DistSet} for some non-empty set $\mathcal X^0_{\vect\theta}$ of initial conditions. 
\end{Definition}

\begin{Definition} \label{Def:Lyap} Let us define the set $\vect \Theta_\mathrm L$ of parameters $\vect\theta$ such that the value function $\hat V_{\vect\theta}(\vect s) $ defined by \eqref{eq:RNMPC} is a Lyapunov function for the dynamics \eqref{eq:StateTraj}-\eqref{eq:DistSet} with respect to a set $\mathbb L_{\vect\theta}$.
\end{Definition}
In order to make Definition~\ref{Def:Lyap} more concrete, we mention as an example the following conditions for $\hat V_{\vect\theta}(\vect s)$ to be a Lyapunov function, which are classic in robust MPC:
\begin{enumerate}
	\item it is lower and upper bounded by $\mathcal K_\infty$ functions;
	\item $\hat V_{\vect\theta}(\vect s_+)  \leq \gamma \hat V_{\vect\theta}(\vect s) + \delta_{\vect\theta}$ holds for all $\vect s\in \mathcal X^0_{\vect\theta}$ and where
	\begin{align*}
		\vect s_+ \in  \left\{\, \vect F_{\vect\vartheta}\left(\vect s, \vect \pi_{\vect\theta}\left(\vect v, \vect s\right), \vect w\right)\,|\,\forall  \,\vect w \in \mathbb W_{\vect\theta} \right\},
	\end{align*}
	for some positive constants $\delta_{\vect\theta}$, and $\gamma<1$.
\end{enumerate}

Then set $\mathbb L_{\vect\theta}$ is defined as:
\begin{align}
\mathbb L_{\vect\theta} = \left\{\,\vect s\,\left|\,  \hat V_{\vect\theta}(\vect s)  \leq \frac{\delta_{\vect\theta}}{1-\gamma}\right.\right\},
\end{align}
and the model \eqref{eq:StateTraj}-\eqref{eq:DistSet} is stabilized to $\mathbb L_{\vect\theta}$ for any initial condition $\vect s \in \mathcal X^0_{\vect\theta}$, see \cite{Rawlings2017}. In Section~\ref{sec:MPCStability}, we will use this definition of Lyapunov function for the sake of simplicity, though our results hold in general, \textit{mutatis mutandis}.

In the context of Definitions \ref{Def:RecFeas} and \ref{Def:Lyap}, the $\sigma$-safety of the real system is guaranteed for $\vect\theta\in\vect\Theta_\mathcal D\cap \vect\Theta_\mathrm F$, and the stability of the real system is guaranteed for $\vect\theta \in\vect\Theta_\mathcal{D}\,\cap\, \vect\Theta_\mathrm{L}$ with probability at least $\sigma$. We note that establishing conditions on $\vect\theta$ such that $\vect\theta \in\vect\Theta_\mathrm{F}\cap \vect\Theta_\mathrm{L}$ is typically done in practice via min-max robust MPC or tube-based MPC \cite{Rawlings2017}. We ought to stress that the conditions above are provided as an example which is fairly general, as it includes the case of~\cite[Section~3.4]{Rawlings2017}, as well as~\cite{Mayne2005,Zanon2021a}. However, more generic conditions are applicable and can be used in our theory.

\begin{Remark} \label{rem:lemma1}
	We observe that Lemma~\ref{lem:safe_stable_policies} is of little practical use since it constructs a specific MPC scheme using  a safe and stabilizing policy which is not known. Nevertheless, it allows us to obtain an important theoretical consideration. Provided that the MPC parametrization is rich enough, there does exist a parameter which yields a non-conservative model \eqref{eq:StateTraj}-\eqref{eq:DistSet}, and a design of the MPC cost and feedback \eqref{eq:Policy} such that the scheme we construct in the proof of the lemma can be captured by an adequate choice of the parameter. This means that there exists a parameter $\vect{\theta}\in \vect\Theta_\mathcal{D}\cap \vect\Theta_\mathrm{F}\cap \vect\Theta_\mathrm{L}$. Consequently, the intersection of these three sets is nonempty for a rich-enough parametrization of the robust MPC scheme.
\end{Remark}

\begin{Assumption} \label{Ass:nicesystem} 
	In the remainder of the paper, we will use the following assumptions:
	\begin{enumerate}
		\item[1.] The set of feasible initial conditions $\mathcal X^0_{\vect\theta}$ is compact and continuous in $\vect\theta$.
		\item[2.] The conditional density \eqref{eq:Transition} underlying the real system is bounded for all state-action pairs, i.e.,
		\begin{align}
			\label{eq:level_set}
			\varphi \left[\,\vect s_+\,|\,\vect s,\vect a\,\right] \leq \bar{\varphi} < \infty,\quad \forall\,\vect s_+,\vect s,\vect a.
		\end{align}
	\end{enumerate}
\end{Assumption}
These technical assumptions will be required in some parts of the theory proposed in this paper. 
We ought to note here that the compactness of set $\mathcal X^0_{\vect\theta}$ is a mild assumption for a well-posed MPC scheme, and is standard in the robust MPC literature \cite{Chisci2001,Mayne2005}. The continuity of the set with respect to $\vect\theta$ holds if, e.g., all functions involved in the MPC scheme are sufficiently differentiable, and if the MPC formulation satisfies standard regularity assumptions for Nonlinear Programs \cite{Nocedal2006,Fiacco1983}. Finally, we stress that one specific setting of learning for robust MPC trivially satisfies this continuity assumption. Indeed, if the learning is not allowed to change the model and safety constraints in the robust MPC scheme, then $\mathcal X^0_{\vect\theta}$ is independent of $\vect{\theta}$ and continuity trivially follows. We should underline here that this case is in principle not restrictive. Indeed, as argued in~\cite{Gros2020}, optimality can be recovered by adjusting the cost only, then such a choice does not introduce any theoretical restriction. Assumption~\ref{Ass:nicesystem}.2 requires that all the states of the real system are subject to some stochasticity. It is a technical assumption aimed at simplifying the subsequent discussions, and could arguably be relaxed at the cost of making the subsequent argumentations significantly more technical.

\subsection{Safety \& Stability Constraints in RL}
In the following, we will consider the safety and stability conditions detailed in this section as constraints applied to the RL steps updating the policy parameters. More specifically, similarly to \cite{Zanon2021}, we consider that the RL steps taken on the robust MPC scheme are feasible steps $\Delta\vect\theta$ taken on the constrained optimization problem: 
\begin{subequations}
	\label{eq:ConstRL}
	\begin{align}
	\min_{\vect\theta}&\quad J\left(\vect\pi_{\vect \theta}\right), \label{eq:ConstRL:Cost} \\
	\mathrm{s.t.}&\quad \vect\theta\in\vect\Theta_\mathrm{L} \cap \vect\Theta_\mathrm{F}\cap \vect\Theta_\mathcal{D},  \label{eq:ConstRL:Const}
	\end{align}
\end{subequations}
in the sense that each parameter update $\vect\theta_{p+1}= \vect\theta_p+\Delta\vect\theta$, labelled by the index $p$, satisfies \eqref{eq:ConstRL:Const} and reduces the cost \eqref{eq:ConstRL:Cost}. More specifically, the RL steps will be computed according to:
\begin{subequations}
	\label{eq:ConstRL:SQP}
	\begin{align}
	\min_{\vect\theta_{p+1}}&\quad \frac{1}{2}\left\|\vect\theta_{p+1}-\vect\theta_p\right\|^2_H +\alpha \nabla_{\vect\theta}J\left(\vect\pi_{\vect \theta_p}\right)^\top\left(\vect\theta_{p+1}-\vect\theta_p\right), \label{eq:ConstRL:Cost:SQP}\\
	\mathrm{s.t.}&\quad \vect\theta_{p+1}\in\vect\Theta_\mathrm{L} \cap \vect\Theta_\mathrm{F}\cap \vect\Theta_\mathcal{D},  \label{eq:ConstRL:Const:SQP}
	\end{align}
\end{subequations}
for some positive-definite matrix $H\approx \nabla^2_{\vect\theta}J\left(\vect\pi_{\vect \theta}\right)$, and some $\alpha\in(0,1]$. We should recall here that for any $H\succ0$ and $\alpha$ small enough, the sequence $\vect\theta_{0,\ldots,\infty}$ stemming from \eqref{eq:ConstRL:SQP} converges to a (possibly local) solution of \eqref{eq:ConstRL} \cite{Nocedal2006}.

It is useful to observe here that classic policy gradient methods \cite{Silver2014,Sutton1999} in RL are typically based on \eqref{eq:ConstRL:Cost:SQP}, with the exclusion of the safety and stability constraint \eqref{eq:ConstRL:Const:SQP}. The identity matrix is often used for $H$ when the policy is based on very high dimensional approximators such as, e.g., Deep Neural Networks.

We will assume here that the gradient $\nabla_{\vect\theta}J$ in \eqref{eq:ConstRL:Cost:SQP} is either evaluated directly via actor-critic or policy search techniques, or replaced by a surrogate based on Q-learning techniques, all formed using data collected on the real system in closed-loop with policy $\vect\pi_{\vect \theta}$. The safety and stability constraints \eqref{eq:ConstRL:Const} will then be built based on \eqref{eq:SafeConst}, \eqref{eq:StateTraj}, \eqref{eq:DistSet},  and \eqref{eq:SafeInclusion}. 
In the remainder of the paper, a mild technical assumption on the Nonlinear Program \eqref{eq:ConstRL:SQP} will be very helpful.
\begin{Assumption} \label{Ass:niceSets} 
	The solution $\vect \theta_{p+1}$ of \eqref{eq:ConstRL:SQP} is continuous with respect to $\alpha$ in a neighborhood of $\alpha =0$.
\end{Assumption}
Assumption \ref{Ass:niceSets} follows from technical assumptions on the set $\vect\Theta_\mathrm{L} \cap \vect\Theta_\mathrm{F}\cap \vect\Theta_\mathcal{D}$, which we propose to not discuss extensively here for the sake of brevity. In particular, we note that Assumption \ref{Ass:niceSets} naturally holds if the set $\vect\Theta_\mathrm{L} \cap \vect\Theta_\mathrm{F}\cap \vect\Theta_\mathcal{D}$ can be represented by a finite set of continuous inequality constraints, and if the resulting problem \eqref{eq:ConstRL:SQP} fulfils classical regularity assumptions and sufficient second-order conditions (SOSC)~\cite{Nocedal2006}.

An important question that needs to be addressed is how feasible parameter updates $\vect\theta_{p+1} = \vect\theta_p+\Delta\vect\theta$ resulting from \eqref{eq:ConstRL:SQP} can be implemented in the robust MPC scheme without jeopardizing the safety and stability of the closed-loop system. We discuss the safety question in the next section using two different approaches. 

\section{Recursive Feasibility with RL-Based Parameter Updates} \label{sec:RecFeas}

Let us consider a sequence of parameters $\vect \theta_{0,\ldots,\infty}$ resulting from \eqref{eq:ConstRL:SQP}, and consider that each parameter $\vect\theta_p$ of that sequence is applied for a certain amount of time (i.e., at least one sampling time of the robust MPC scheme). This section provides conditions such that this sequence of parameter updates does not jeopardize the safety of the corresponding sequence of policies $\vect\pi_{\vect\theta_{0,\ldots,\infty}}$ resulting from the corresponding robust MPC schemes. Note that a parameter update $\vect\theta_p\rightarrow \vect\theta_{p+1}$ occurring at a time sample $i$ means here that $\vect a_i = \vect\pi_{\vect\theta_{p+1}}\left(\vect s_i\right)$ and the inputs $\vect a_j = \vect\pi_{\vect\theta_{p}}\left(\vect s_j\right)$ with $j<i$ are used from the previous parameter update. The next theorem provides a first set of conditions for ensuring the safety of the parameter updates.
\begin{theorem} \label{Th:Recursivity} 
	Assume that for all $p$, parameter $\vect\theta_p$ satisfies $\vect\theta_p\in\vect\Theta_\mathrm F\cap \vect\Theta_\mathcal{D}$, and that the initial conditions $\vect s_0$ are in the set $\mathcal X_{\vect\theta_0}^0$. If each parameter update $\vect\theta_p\rightarrow \vect\theta_{p+1}$ takes place in a state $\vect s_i$ such that 
	\begin{align}
	\label{eq:FeasibleTransition}
	\vect s_i\in\mathcal X_{\vect\theta_{p+1}}^0
	\end{align}
	holds, then the closed-loop trajectories $\vect s_{0,\ldots,\infty}$ resulting from applying the sequence of policies $\vect\pi_{\vect\theta_{0,\ldots,\infty}}$ is $\sigma$-safe.
\end{theorem}
\begin{pf} 
	Assuming that \eqref{eq:IncludedSupport} holds, then a standard result for robust MPC is that if $\vect\theta_{p+1}\in\vect\Theta_\mathrm F$ and if the initial state at which policy $\vect\pi_{\vect\theta_{p+1}}$ is deployed satisfies condition  \eqref{eq:FeasibleTransition}, then policy $\vect\pi_{\vect\theta_{p+1}}$ ensures that the state trajectories are feasible at all time with unitary probability. We then observe that if every parameter update $\vect\theta_p\rightarrow \vect\theta_{p+1}$ is applied under condition  \eqref{eq:FeasibleTransition}, then each policy keeps the state trajectories feasible. As a result, if \eqref{eq:FeasibleTransition} is ensured at every parameter update $\vect\theta_p\rightarrow \vect\theta_{p+1}$, then the entire state trajectory $\vect s_{0,\ldots,\infty}$ resulting from the policy sequence $\vect\pi_{\vect\theta_{0,\ldots,\infty}}$ remains feasible at all time. 
	
	If statement \eqref{eq:IncludedSupport} does not hold, then the closed-loop trajectories $\vect s_{0,\ldots,\infty}$ may become infeasible, though not necessarily. Hence if statement \eqref{eq:IncludedSupport} holds with a probability $\sigma$, then the closed-loop trajectories $\vect s_{0,\ldots,\infty}$ have probability no smaller than $\sigma$ to be feasible. 	$\hfill\square$
\end{pf}

The results of Theorem \ref{Th:Recursivity} can be leveraged in practice by solving the robust MPC schemes associated to both $\vect\theta_{p}$ and $\vect\theta_{p+1}$ in parallel at every sampling instant and selecting the control input associated to $\vect\theta_{p+1}$ as soon as the MPC scheme associated to $\vect\theta_{p+1}$ is feasible. 

The theorem ensures the recursive feasibility of the sequence of robust MPC schemes such that the closed-loop state trajectories are contained in the set $\mathcal X$, within the framework presented in Section \ref{sec:SafeRL}. An important caveat, though, is that there is no guarantee that condition \eqref{eq:FeasibleTransition} can be met in finite time by the closed-loop trajectories under policy $\vect\pi_{\vect\theta_p}$. As a result, it might be possible that a parameter update $\vect\theta_{p+1}$, though feasible for  \eqref{eq:ConstRL}, yields an update condition \eqref{eq:FeasibleTransition} that never becomes satisfied, hence blocking the learning process. The remainder of this section proposes two different approaches to tackle that issue, using either backtracking or additional constraints in \eqref{eq:ConstRL}.

In order to support and simplify the coming argumentation, it is useful to introduce a technical lemma. Let us consider the trivial locally compact measure on $\mathbb R^n$, associating to any compact set $\mathbb A\subset \mathbb R^n$ the bounded positive real number:
\begin{align}
\mu\left(\mathbb A\right) = \int_{\mathbb A} \mathrm d\vect s.
\end{align}
\begin{Lemma} \label{Lem:CoolLemma}
	Suppose that Assumption~\ref{Ass:nicesystem}.2 holds and consider
	an arbitrary policy $\vect a = \vect\pi\left(\vect s\right)$, yielding a (continuous) Markov Chain $\vect s_{0,\ldots,\infty}$. Then for any set $\mathbb A$ and initial condition $\vect s_0\in\mathbb A$, the following inequality holds:
	\begin{align}
	\label{eq:Lemma1}
\mathbb P[\,\vect s_{0,\ldots,\infty} \in \mathbb A\,] \leq \mu\left(\mathbb A\right)\bar{\varphi}
\end{align}
where $\bar{\varphi}$ is defined by \eqref{eq:level_set}.
\end{Lemma}
\begin{pf} 
	Denoting $\phi[\vect s_k]$ the density of the Markov chain at time $k$, we observe that:
	\begin{align}
	\phi[\vect s_k] &:= \int \varphi\left[\,\vect s_k\,|\,\vect s_{k-1},\vect \pi\left(\vect s_{k-1}\right)\,\right]  \phi[\vect s_{k-1}] \mathrm d\vect s_{k-1}\leq  \bar{\varphi} 
	\end{align}
	holds for any $k>0$. It follows that
	\begin{align}
	\mathbb P[\vect s_k\in\mathbb A] = \int_{\mathbb A} \phi[\vect s_k]\mathrm d\vect s_k \leq \mu\left(\mathbb A\right) \bar{\varphi} .
	\end{align}
	Then \eqref{eq:Lemma1} follows from the Fr\'echet inequalities, stating:
	\begin{align}
	\mathbb P[\vect s_{0,\ldots,\infty} \in \mathbb A] \leq \inf_k\,\,\mathbb P[\vect s_k \in \mathbb A]  \leq  \mu\left(\mathbb A\right)\bar{\varphi}.
	\end{align}
	$\hfill\square$

\end{pf}

\subsection{Parameter Update via Backtracking}
In this subsection, we consider the use of backtracking on the parameter updates computed according to \eqref{eq:ConstRL:SQP} to ensure the feasibility of updating the parameters in finite time. For the sake of simplicity in the following developments, rather than a line-search strategy \cite{Nocedal2006}, we use a gradient adaptation strategy in the cost \eqref{eq:ConstRL:Cost:SQP}, by iteratively reducing parameter $\alpha$, therefore generating a step ranging from a full step ($\alpha = 1$) to $\Delta\vect\theta=0$ (with $\alpha = 0$). The following theorem then guarantees that there is some $\alpha >0$ such that the probability that the parameter update condition \eqref{eq:FeasibleTransition} is not met in finite time is less than $1-\sigma$.

\begin{theorem} \label{eq:AwesomeBacktrack} 
	Consider the closed loop trajectory $\vect s_{i,\ldots,\infty}$ under policy $\vect\pi_{\vect\theta_p}$ starting at the physical sampling time $i$ with the initial state $\vect s_i$. Consider the parameter update $\vect\theta_{p+1}(\alpha)$ (where we highlight the dependency on $\alpha$) resulting from \eqref{eq:ConstRL:SQP} and suppose that Assumptions~\ref{Ass:nicesystem}-\ref{Ass:niceSets} hold. Then the probability that the update condition \eqref{eq:FeasibleTransition} is not met in finite time can be made arbitrarily small by selecting $\alpha$ small enough, i.e., the following limit holds:
	\begin{align}
	\label{eq:LimitProbability}
	\lim_{\alpha \rightarrow 0}\mathbb P\left[\vect s_{i,\ldots,\infty}\notin\mathcal X_{\vect\theta_{p+1}(\alpha)}^0\right] \leq 1-\sigma.
	\end{align}
\end{theorem}
A simple interpretation of Theorem \ref{eq:AwesomeBacktrack} is that it is always possible to backtrack to a short-enough parameter update ($\alpha$ small enough) such that the update condition \eqref{eq:FeasibleTransition} becomes satisfied in finite time with probability arbitrarily close to $1-\sigma$.

The intuition behind this result is that, by continuity arguments, $\mathcal X_{\vect\theta_{p+1}(\alpha)}^0$ tends to $ \mathcal X_{\vect\theta_{p}}^0$ as $\alpha$ becomes small, such that the two sets match asymptotically. Moreover, we observe that under policy $\vect\pi_{\vect\theta_p}$, the closed-loop trajectories evolve in set $\mathcal X_{\vect\theta_{p}}^0$ and the update is infeasible if they are outside of set $\mathcal X_{\vect\theta_{p+1}(\alpha)}^0$. It follows that for a parameter update to be blocked forever despite $\alpha$ being arbitrarily small, if \eqref{eq:IncludedSupport} holds, the closed-loop state trajectories under policy $\vect\pi_{\vect\theta_p}$ need to evolve on an infinitely small set. This would require unbounded densities in the real closed-loop dynamics \eqref{eq:Transition}, which is excluded by Assumption \ref{Ass:nicesystem}, or that \eqref{eq:IncludedSupport} does not hold. We formalize these explanations in the next proof.

\begin{pf} (of Theorem \ref{eq:AwesomeBacktrack}) 
	If \eqref{eq:IncludedSupport} holds, we first observe that $\vect\theta_{p+1}(0) = \vect\theta_{p}$ trivially holds, such that 
	\begin{align}
	\vect s_{i,\ldots,\infty}\in\mathcal X_{\vect\theta_{p+1}(0)}^0
	\end{align}
	holds by construction. Let us further define the set:
	\begin{align}
	\Delta\mathcal X^0(\vect\theta_{p},\vect\theta_{p+1}) = \left\{\,\vect s \,\left|\, \vect s \in \mathcal X^0_{\vect\theta_p} \quad \text{and}\quad \vect s\notin \mathcal X^0_{\vect\theta_{p+1}}\,\right.\right\},
	\end{align}
	such that $\Delta\mathcal X^0(\vect\theta_{p},\vect\theta_{p}) = \emptyset$. A trajectory $\vect s_{k,\ldots,\infty}$ in closed-loop under policy $\vect\pi_{\vect\theta_p}$ that never satisfies the update condition \eqref{eq:FeasibleTransition} must evolve in $\Delta\mathcal X^0(\vect\theta_{p},\vect\theta_{p+1})$. We observe that by Assumption \ref{Ass:niceSets} $\vect\theta_{p+1}(\alpha)$ is continuous in a neighborhood of $\alpha=0$, and by Assumption \ref{Ass:nicesystem}.1 we have that the set $\Delta\mathcal X^0(\vect\theta_{p},\vect\theta_{p+1})$ is continuous in $\vect\theta_{p+1}$. It follows that 
	\begin{align}
	\label{eq:ZeroMeasure}
	\lim_{\alpha\rightarrow 0}\, \mu\left(\Delta\mathcal X^0(\vect\theta_{p},\vect\theta_{p+1}(\alpha))\right) = 0.
	\end{align}
	We can then conclude using Lemma \ref{Lem:CoolLemma}:
	\begin{align}
	\label{eq:NoLimitProbability}
	&\lim_{\alpha\rightarrow 0}\mathbb P\left[\vect s_{i,\ldots,\infty}\notin\mathcal X_{\vect\theta_{p+1}(\alpha)}^0\right] = \\&\qquad \lim_{\alpha\rightarrow 0}\mathbb P\left[\vect s_{i,\ldots,\infty}\in\Delta\mathcal X^0(\vect\theta_{p},\vect\theta_{p+1}(\alpha))\right] =0.\nonumber
	\end{align}
	Since \eqref{eq:IncludedSupport} holds with probability $\sigma$, \eqref{eq:LimitProbability} readily follows.
	$\phantom{.}$$\hfill\square$
\end{pf}
A practical implementation of the backtracking approach is detailed in Algorithm~\ref{Alg:SafeRLBacktrack}. The implementation consists in reducing parameter $\alpha$ if the update condition is not met for $n$ time steps. We ought to stress here that lines 3 and 14 of Algorithm~\ref{Alg:SafeRLBacktrack} can be performed offline independently of the state of the system and of the robust MPC schemes. It follows that the online computational burden is limited to solving two independent robust MPC schemes, possibly in parallel.

\IncMargin{1.5em}
\begin{algorithm2e}[t]
	\caption{Safe and Stable learning - backtracking}
	\label{Alg:SafeRLBacktrack}
	\SetKwInOut{Input}{Input}
	\Input{MPC parameter $\vect \theta$, and $\varrho$, $n$, $H$}
	\While{$\mathrm{Learning}$}{
		Set $\mathrm{update} = \mathrm{false}$, $\alpha = 1$, $\mathrm{fail} = 0$ \\
		Compute $\vect\theta_{+}$ from \eqref{eq:ConstRL:SQP} \\
		\While{not $\mathrm{update}$ }{
			Compute MPC solution $\vect u_0$ from $\vect\theta$ \\
			Compute MPC solution $\vect u_0^+$ from $\vect\theta_{+}$ \\
			\If{MPC solution from $\vect\theta_{+}$ is feasible}
			{
				Set $\vect u_0\leftarrow \vect u_0^+$ and $\vect\theta\leftarrow \vect\theta_+$   \\
				$\mathrm{update} = \mathrm{true}$
			}
			\Else
			{
				$\mathrm{fail} = \mathrm{fail}+1$
			}
			Apply input $\vect u_0$ to the system \\
			\If{$\mathrm{fail} \geq n$}{
				$\alpha = \varrho \alpha$ and recompute $\vect\theta_{+}$ from \eqref{eq:ConstRL:SQP} \\
			}
		}
	}
\end{algorithm2e} 

\subsection{Parameter Updates via Constrained Feasibility}

As an alternative to backtracking, we propose next an approach imposing additional constraints in \eqref{eq:ConstRL:SQP}.  We then no longer rely on taking short-enough steps in $\vect\theta$ to achieve the feasibility of the parameter updates, but rather form an update that is feasible by construction. This entails that updates can be performed at every time step $i$ such that $p=i$, hence we will use the notation $\vect{\theta}_i$ throughout this section.

We define as $\mathbb X_1^{\vect{\theta}_{i+1}}(\vect{s}_i,\vect{\pi}_{\vect{\theta}_i}(\vect{s}_i))$ the 1-step dispersion set starting from state $\vect{s}_i$, applying the input $\vect{a}=\vect{\pi}_{\vect{\theta}_i}(\vect{s}_i)$, and using set $\mathbb{W}_{\vect{\theta}_{i+1}}$ to model the stochasticity of the system. Note the subtle but important difference with $\mathbb X_1^{\vect{\theta}_{i+1}} = \mathbb X_1^{\vect{\theta}_{i+1}}(\vect{s}_i,\vect{\pi}_{\vect{\theta}_{i+1}}(\vect{s}_i))$, where we apply action $\vect{a}=\vect{\pi}_{\vect{\theta}_{i+1}}(\vect{s}_i)$ instead of $\vect{a}=\vect{\pi}_{\vect{\theta}_i}(\vect{s}_i)$.

\begin{theorem} \label{Th:TrajectoryFeasibility:Unsynch}
	The parameter update $\vect\theta_{i+1}$ given by
	\begin{subequations}
		\label{eq:ConstRL:FeasibleUpdates}
		\begin{align}
		\min_{\vect\theta_{i+1}}&\quad \frac{1}{2}\left\|\vect\theta_{i+1}-\vect\theta_i\right\|^2_H +\nabla_{\vect\theta}J\left(\vect\pi_{\vect \theta_i}\right)^\top\left(\vect\theta_{i+1}-\vect\theta_i\right), \label{eq:ConstRL:FeasibleUpdates:Cost} \\
		\mathrm{s.t.}&\quad \vect\theta_{i+1}\in\vect\Theta_\mathrm{L} \cap \vect\Theta_\mathrm{F}\cap \vect\Theta_\mathcal{D},  \label{eq:ConstRL:FeasibleUpdates:Const} \\
		&\quad \mathcal{X}^0_{\vect \theta_{i+1}}\supseteq \mathbb X_1^{\vect{\theta}_{i+1}}(\vect{s}_i,\vect{\pi}_{\vect{\theta}_i}(\vect{s}_i)).
		\label{eq:FeasibleUpdateConstraint}
		\end{align}
	\end{subequations}
	satisfies \eqref{eq:FeasibleTransition} by construction with probability at least $\sigma$.
\end{theorem}
\begin{pf}
	Equation~\eqref{eq:IncludedSupport} entails $\vect{s}_{i+1} \in \mathbb X_1^{\vect{\theta}_{i+1}}(\vect{s}_i,\vect{\pi}_{\vect{\theta}_i}(\vect{s}_i))$. 
	Using~\eqref{eq:FeasibleUpdateConstraint}, 
	$\mathbb X_1^{\vect{\theta}_{i+1}}(\vect{s}_i,\vect{\pi}_{\vect{\theta}_i}(\vect{s}_i))\subseteq \mathcal{X}^0_{\vect{\theta}_{i+1}}$ holds, and robust MPC is feasible for all possible realizations of $\vect{s}_{i+1}$, i.e., $\vect{s}_{i+1}\in \mathcal{X}^0_{\vect{\theta}_{i+1}}$. Since~\eqref{eq:IncludedSupport} holds with probability $\sigma$, \eqref{eq:FeasibleTransition} holds with probability at least $\sigma$.
	$\hfill\square$
\end{pf}

We elaborate next on how constraint~\eqref{eq:FeasibleUpdateConstraint} can be formulated. 
We observe that recursive feasibility of MPC~\eqref{eq:RNMPC} implies that, if a given state is feasible, then the tube 
around the predicted trajectory is also feasible, such that
\begin{align}
	\label{eq:equivalentSetInclusion}
	\mathbb X_1^{\vect{\theta}_{i+1}}(\vect{s}_i,\vect{\pi}_{\vect{\theta}_i}(\vect{s}_i))\subseteq \mathcal{X}^0_{\vect{\theta}_{i+1}} && \Leftarrow && \vect{s}_i \in \mathcal{X}^0_{\vect{\theta}_{i+1}}(\vect{\pi}_{\vect{\theta}_i}(\vect{s}_i)),
\end{align}
where $\mathcal{X}^0_{\vect{\theta}_{i+1}}(\vect{a})$ defines the set of states $\vect{s}$ for which the robust MPC scheme~\eqref{eq:RNMPC} is feasible under the additional constraint $\vect{\bar u}_0=\vect{a}$. 
The second condition in~\eqref{eq:equivalentSetInclusion} is more easily written as a condition on the parameters and the nominal MPC trajectory: a detailed discussion on how this is done is provided in~\cite{Zanon2021} for linear tube MPC. The main difference between that approach and the one used in this paper is that in that case the constraint takes the form $\vect{h}(\vect{v},\vect{\theta}) \leq 0$, while in this paper it takes the form $\vect{h}(\vect{v}^\star(\vect{\theta}),\vect{\theta}) \leq 0$. Since the main idea is unchanged, we do not provide further details for the sake of brevity.

We prove next that constraint~\eqref{eq:ConstRL:FeasibleUpdates} is non-blocking, i.e., that the parameter update yielded by~\eqref{eq:ConstRL:FeasibleUpdates} cannot be $\vect{\theta}_{i+1}=\vect{\theta}_i$ at all times, unless $\vect{\theta}_i=\vect{\theta}_\star$. Note that Theorem~\ref{Th:TrajectoryFeasibility:Unsynch} does not require Assumption~\ref{Ass:nicesystem} nor Assumption~\ref{Ass:niceSets}. However, both assumptions are needed in order to be able to prove the non-blocking property.
\begin{theorem}
	Consider the closed loop trajectory $\vect s_{i,\ldots,\infty}$ under policy $\vect\pi_{\vect\theta_i}$ starting at the physical sampling time $i$ with the initial state $\vect s_i$. Suppose that Assumptions~\ref{Ass:nicesystem}-\ref{Ass:niceSets} hold.
	Assume further that Assumption~\ref{Ass:nicesystem}.1 holds also for $\mathcal{X}^0_{\vect{\theta}_{i+1}}(\vect{a})$, i.e., if the first action is fixed in the MPC scheme, the set of feasible initial conditions is compact and continuous in~$\vect{\theta}$. 
	Then, the probability that~\eqref{eq:ConstRL:FeasibleUpdates} yields $\vect{\theta}_{i+1}=\vect{\theta}_i\neq \vect{\theta}_\star$ for all times is zero, i.e.,
	\begin{align}
		\label{eq:infeas_update}
		\mathbb{P}\left [\vect{\theta}_{p+1}=\vect{\theta}_p\neq \vect{\theta}_\star, \ p=i,\ldots,\infty \right ] =0.
	\end{align}
\end{theorem}
\begin{pf}
	In order to prove the result, we will prove that there does exist a parameter update $\vect{\theta}_{i+1}\neq\vect{\theta}_i$ which does decrease the cost~\eqref{eq:ConstRL:FeasibleUpdates:Cost} and satisfy~\eqref{eq:FeasibleUpdateConstraint}. To that end, we consider $\vect\theta_{i+1}(\alpha)$ resulting from \eqref{eq:ConstRL:SQP}, and we prove next that 
	\begin{align}
		\label{eq:infeas_update2}
		&\hspace{-1em}\lim_{\alpha\to0} \mathbb{P}\left [\mathcal{X}^0_{\vect \theta_{i+1}(\alpha)} \not\supseteq \mathbb X_1(\vect{s}_j,\vect{\pi}_{\vect{\theta}_i}(\vect{s}_j)), \ j=i,\ldots,\infty \right ]  \\
		&\hspace{4em}\leq\lim_{\alpha\rightarrow 0}\mathbb P\left[\vect s_{i,\ldots,\infty}\notin\mathcal X_{\vect\theta_{i+1}(\alpha)}^0(\vect{\pi}_{\vect{\theta}_i}(\vect{s}_i)))\right] = 0, \nonumber
	\end{align}
	where we used~\eqref{eq:equivalentSetInclusion} to obtain the inequality above. Because $H\succ0$, any update $\vect{\theta}_{i+1}(\alpha)$ reducing~\eqref{eq:ConstRL:Cost:SQP} must also be a descent direction for~\eqref{eq:ConstRL:FeasibleUpdates:Cost}. Consequently, if additionally $\vect{\theta}_{i+1}(\alpha)$ is feasible for~\eqref{eq:FeasibleUpdateConstraint}, then~\eqref{eq:ConstRL:FeasibleUpdates} cannot yield a $0$ update.
	
	Since by using $\alpha=0$,~\eqref{eq:ConstRL:SQP} yields $\vect\theta_{i+1}(0) = \vect\theta_{i}$,  we have
	\begin{align*}
		\vect s_{i,\ldots,\infty}\in\mathcal X_{\vect\theta_{i+1}(0)}^0(\vect{\pi}_{\vect{\theta}_i}(\vect{s}_i)) = \mathcal X_{\vect\theta_{i}}^0(\vect{\pi}_{\vect{\theta}_i}(\vect{s}_i)) = \mathcal X_{\vect\theta_{i}}^0.
	\end{align*}
	We use~\eqref{eq:equivalentSetInclusion} to define the set:
	\begin{align}
	&\Delta\mathcal X^0_{\vect{\pi}_{\vect{\theta}_i}(\vect{s}_i)}(\vect\theta_{i},\vect\theta_{i+1})  \\
	&\hspace{4em}= \left\{\,\vect s \,\left|\, \vect s \in \mathcal X^0_{\vect\theta_i}, \ \text{and}\ \vect s\notin \mathcal X^0_{\vect\theta_{i+1}}(\vect{\pi}_{\vect{\theta}_i}(\vect{s}_i))\,\right.\right\}, \nonumber
	\end{align}
	i.e., the set for which $\vect\theta_{i+1}$ violates the constraint~\eqref{eq:FeasibleUpdateConstraint}.
	Note that $\Delta\mathcal X^0_{\vect{\pi}_{\vect{\theta}_i}(\vect{s}_i)}(\vect\theta_{i},\vect\theta_{i}) = \emptyset$. Consider a trajectory $\vect s_{i,\ldots,\infty}$ in closed-loop under policy $\vect\pi_{\vect\theta_i}$ such that parameter $\vect{\theta}_{i+1}(\alpha)$ solves~\eqref{eq:ConstRL:SQP} but never satisfies constraint \eqref{eq:FeasibleUpdateConstraint}.
	By definition such trajectory must evolve in set $\Delta\mathcal X^0_{\vect{\pi}_{\vect{\theta}_i}(\vect{s}_i)}(\vect\theta_{i},\vect\theta_{i+1}(\alpha))$. We observe that, by Assumption~\ref{Ass:niceSets}, $\vect\theta_{i+1}(\alpha)$ is continuous in a neighborhood of $\alpha=0$, and by assumption we have that the set $\Delta\mathcal X^0_{\vect{\pi}_{\vect{\theta}_i}(\vect{s}_i)}(\vect\theta_{i},\vect\theta_{i+1})$ is continuous in $\vect\theta_{i+1}$. It follows that 
	\begin{align}
	\label{eq:ZeroMeasure2}
	\lim_{\alpha\rightarrow 0}\, \mu\left(\Delta\mathcal X^0_{\vect{\pi}_{\vect{\theta}_i}(\vect{s}_i)}(\vect\theta_{i},\vect\theta_{i+1}(\alpha))\right) = 0.
	\end{align}
	We can then conclude using Lemma \ref{Lem:CoolLemma}:
	\begin{align}
	\label{eq:NoLimitProbability2}
	&\lim_{\alpha\rightarrow 0}\mathbb P\left[\vect s_{i,\ldots,\infty}\notin\mathcal X_{\vect\theta_{i+1}(\alpha)}^0(\vect{\pi}_{\vect{\theta}_i}(\vect{s}_i))\right] = \\&\qquad \lim_{\alpha\rightarrow 0}\mathbb P\left[\vect s_{i,\ldots,\infty}\in\Delta\mathcal X^0_{\vect{\pi}_{\vect{\theta}_i}(\vect{s}_i)}(\vect\theta_{i},\vect\theta_{i+1}(\alpha))\right] =0.\nonumber
	\end{align}
	$\phantom{.}$$\hfill\square$
\end{pf}

\section{Stability of MPC with Parameter Updates} \label{sec:MPCStability}
In the previous section, we investigated the recursive feasibility of performing RL-based parameter updates on the MPC scheme. In this section we will discuss the stability of a sequence of MPC schemes satisfying the recursive feasibility conditions discussed above. We will first discuss the joint stability of the state $\vect{s}$ and parameters $\vect{\theta}$; and then relax our assumptions, treat the updates of $\vect{\theta}$ as  a perturbation acting on the system, and prove Input-to-State Stability (ISS).

In order to discuss joint state and parameter stability, we will show that, assuming that the sequence of parameters converges linearly, the sequence of MPC policies stabilizes the system in the state-parameter space. 
If $\vect\theta_p\in\vect\Theta_\mathrm L \cap \vect\Theta_\mathrm F\cap \vect\Theta_\mathcal{D}$ for all $ p$, the sequence of parameters $\vect \theta_{0,\ldots,\infty}$ yields a sequence of Lyapunov functions $\hat V_{\vect\theta_p}$ on their respective feasible sets $\mathcal{X}^0_{\vect \theta_p}$. Hence each MPC with parameter $\vect\theta_p$ is stabilizing the system trajectory to the corresponding level set $\mathbb{L}_{\vect{\theta}_p}$.
The stability of the system trajectories when updating the parameters $\vect\theta_p$ can then be investigated by piecing together the individual Lyapunov functions $\hat V_{\vect \theta_{0,\ldots,\infty}}$, and by assuming some regularity condition on the functions $\hat V_{\vect \theta_{0,\ldots,\infty}}$ as well as a sufficiently fast convergence of the parameter sequence $\vect \theta_{0,\ldots,\infty}$. This statement is formalized in the next theorem, for which we need the two following assumptions.  

\begin{Assumption}
	\label{ass:convergence}
	 At every parameter update $\vect\theta_p\rightarrow\vect\theta_{p+1}$, the inequality:
	\begin{align}
		\label{eq:ThetaConvergence}
		\left\|\vect\theta_{p+1} - \vect\theta_{\star}\right\| \leq r\left\|\vect\theta_{p} - \vect\theta_{\star}\right\| 
	\end{align}
	holds for some $r \in]0,1[$, where $\vect\theta_{\star}$ is the solution of \eqref{eq:ConstRL}.
\end{Assumption}
\begin{Assumption}
	\label{ass:regularity}
	The condition
	\begin{align}
		\label{eq:VhatRegularity}
		\sup_{\vect s\in \mathcal X^0_{\vect\theta_p}\cap\,  \mathcal X^0_{\vect\theta_{p+1}}}\left|\hat V_{\vect\theta_{p+1}}\left(\vect s\right) -\hat V_{\vect\theta_p}\left(\vect s\right) \right| \leq \alpha_V\left\|\vect\theta_{p+1} - \vect\theta_{p}\right\|
	\end{align}
	holds for all $p$ for some $ \alpha_V>0$.
\end{Assumption}

Before providing the result, let us discuss these two assumptions. Assumption~\ref{ass:convergence} essentially requires that the parameter updates converge at a linear rate. This type of convergence holds for exact gradient methods, i.e., if the policy gradient $\nabla_{\vect\theta} J(\vect \pi_{\vect\theta} )$ used to compute the parameter updates in, e.g., \eqref{eq:ConstRL:SQP} is evaluated exactly. We ought to stress here that in a learning context, the gradient is typically evaluated from data and therefore is stochastic by nature. The amount of stochastic perturbation around the average value asymptotically decays as the amount of data used in the gradient evaluation increases. As a result, Assumption~\ref{ass:convergence} is to be taken as asymptotically valid, as well as the subsequent result presented in Theorem \ref{Th:TrajectoryStability}. A stability result requiring an assumption weaker than Assumption~\ref{ass:convergence} is presented later in the text. That latter result is, however, weaker.

In order to discuss the conservatism of the regularity Assumption~\ref{ass:regularity}, we observe first that in general both the policy and the value function can be discontinuous~\cite{Rawlings2017}. Indeed, the continuity of the value function of optimization problems is known to be intricate, with the notable exception of linear robust MPC with a quadratic cost and polyhedral uncertainty. Assuming that the parameter sequence $\vect\theta_{0,\ldots,\infty}$ belongs entirely to a bounded and connected (BC) set $\Theta$ such that $\mathcal X^0_{\vect\theta}$ is non-empty everywhere in $\Theta$, we observe that \eqref{eq:VhatRegularity} holds if $\hat V_{\vect\theta}$ is Lipschitz continuous on $\Theta$ with a bounded Lipschitz constant for all $\vect s$ in the applicable domain of definition. This, in turn, holds if for all $\vect s$ in the domain of definition, $\hat V_{\vect\theta}$ is almost everywhere differentiable with respect to $\vect\theta$ on $\Theta$, with bounded and Lebesgue integrable derivatives. Since in the general case discontinuities cannot be excluded, one might be tempted to conclude that~\eqref{eq:VhatRegularity} is in general violated. However, one should consider that, under very mild regularity assumptions, such discontinuities can only occur on a set of zero measure, i.e., condition~\eqref{eq:VhatRegularity} holds for almost all $\vect{s}$. Indeed, whenever the optimal MPC solution satisfies the strong second order sufficient conditions for optimality, a suitable constraint qualification, and strict complementarity holds, then both the policy and the value function are continuous and differentiable with respect to the parameter $\vect{\theta}$~\cite{Nocedal2006,Fiacco1983}. In case strict complementarity does not hold, then directional derivatives do exist and continuity is not lost, such that~\eqref{eq:VhatRegularity} holds. 
A typical situation in which the policy and eventually also the value function can become discontinuous is when one of the two other conditions is not met. We observe that essentially all NMPC solvers require these two conditions to hold, such that one can conclude that any state-parameter combination which does not pose issues for NMPC solvers satisfies~\eqref{eq:VhatRegularity}. Since the set of problematic points is of zero measure for well-posed problems, the probability of visiting such a state-parameter pair is zero for non-pathological systems~\eqref{eq:Transition}.

\begin{theorem} \label{Th:TrajectoryStability}
	Suppose that Assumptions~\ref{ass:convergence}-\ref{ass:regularity} hold. Let us additionally assume that each parameter is given by \eqref{eq:ConstRL:SQP} or \eqref{eq:ConstRL:FeasibleUpdates} such that $\vect\theta_{p}\in\vect\Theta_\mathrm{L} \cap \vect\Theta_\mathrm{F}\cap \vect\Theta_\mathcal{D}$, and the parameter updates satisfy the parameter update conditions \eqref{eq:FeasibleTransition} or~\eqref{eq:FeasibleUpdateConstraint}. Then the sequence of robust MPC schemes with parameters $\vect\theta_{0,\ldots,\infty}$ is asymptotically stable in the joint state-parameter update space for any $\vect s_0 \in \mathbb X^0_{\vect\theta_0}$, and steers the system trajectory to the level set $\mathbb{L}_{\vect{\theta}_\infty}$. 
\end{theorem}
\begin{pf}
	Consider an augmentation of the state $\vect s$ with the current parameters $\vect\theta_p$. Let us label the augmented state $\vect S$. We then propose the candidate Lyapunov function:
	\begin{align}\label{eq:LyapCandidate}
	W(\vect S) = \hat V_{\vect\theta_p}\left(\vect s\right) + \zeta\Delta_{p}
	\end{align}
	where we label $\Delta_p := \left\|\vect\theta_{p} - \vect\theta_{\star}\right\|$, and $\zeta$ is a positive constant. Function \eqref{eq:LyapCandidate} tackles the regular state space of the system jointly with the parameter update space, and will allow us to establish stability in that joint space for $\zeta$ large enough. We first observe that since $\hat V_{\vect\theta_{0,\ldots,\infty}}\left(\vect s\right)$ are Lyapunov functions, for any $\vect\theta_p$, $W(\vect S)$ is a Lyapunov function in between parameter updates, i.e., $W$ is decreasing along the system trajectory. Moreover, since $\Delta_{p}$ is a norm in the state-parameter space, $W\left(\vect S\right)$ is adequately lower and upper bounded if $\hat V_{\vect\theta_p}$ is. Finally, since by assumption $\Delta_p\rightarrow 0$, the system state $\vect s$ is eventually steered towards $\mathbb{L}_{\vect{\theta}_\infty}$. We observe that between parameter updates the decrease of $W(\vect S)$ under a specific parameter $\vect\theta_p$ holds from:
	\begin{align}
	W(\vect S_{i+1}) &= \hat V_{\vect\theta_p}\left(\vect s_{i+1}\right) + \zeta\Delta_{p} \\
	& \leq \gamma\hat V_{\vect\theta_p}\left(\vect s_{i}\right) +\delta_{\vect\theta_p}+ \zeta\Delta_{p} < W(\vect S_{i}),
	\end{align}
	for any $\vect s_i$ outside of $\mathbb{L}_{\vect{\theta}_p}$. 
	
	Upon updating the parameter $\vect\theta_p\rightarrow \vect\theta_{p+1}$ at a specific time instant $i$, we observe that:
	\begin{align}
	W(\vect S_{i+1}) - W(\vect S_i) = &\, \hat V_{\vect\theta_{p+1}}\left(\vect s_{i+1}\right) - \hat V_{\vect\theta_{p}}\left(\vect s_i\right) \label{eq:Lyap_decrease}\\&+ \zeta\left(\Delta_{p+1} - \Delta_{p}\right). \nonumber
	\end{align}
	If the state at time $\vect s_i$ lies outside of the level set $\mathbb{L}_{\vect{\theta}_{p+1}}$ such that:
	\begin{align}
	\hat V_{\vect\theta_{p+1}}\left(\vect s_{i+1}\right) < \hat V_{\vect\theta_{p+1}}\left(\vect s_{i}\right) 
	\end{align}
	holds under the MPC with parameter $\vect\theta_{p+1}$, then $W$ is decreasing over the parameter updates at time $i$ if:
	\begin{align}
	\hat V_{\vect\theta_{p+1}}\left(\vect s_{i+1}\right) - \hat V_{\vect\theta_{p}}\left(\vect s_i\right) + \zeta\left(\Delta_{p+1} - \Delta_{p}\right) &< \nonumber \\
	\hat V_{\vect\theta_{p+1}}\left(\vect s_{i}\right)  - \hat V_{\vect\theta_{p}}\left(\vect s_i\right) + \zeta\left(\Delta_{p+1} - \Delta_{p}\right) &\leq 0.\label{eq:CondWDecayOverUpdates}
	\end{align}
	Using \eqref{eq:VhatRegularity}, condition \eqref{eq:CondWDecayOverUpdates} holds if:
	\begin{align} \label{eq:OverUpdateDecayCondition}
	\hat V_{\vect\theta_{p+1}}\left(\vect s_{i}\right)  - \hat V_{\vect\theta_{p}}\left(\vect s_i\right) + \zeta\left(\Delta_{p+1} - \Delta_{p}\right) &\leq \nonumber\\
	\alpha_V\left\|\vect\theta_{p+1} - \vect\theta_{p}\right\| + \zeta\left(\left\|\vect\theta_{p+1} - \vect\theta_{\star}\right\| - \left\|\vect\theta_{p} - \vect\theta_{\star}\right\|\right)& \leq 0.
	\end{align}
	Using \eqref{eq:ThetaConvergence} we observe that
	\begin{align}
	\left\|\vect\theta_{p+1} - \vect\theta_{p}\right\| &\leq \left\|\vect\theta_{p+1} - \vect\theta_\star\| + \|\vect\theta_{p} -  \vect\theta_{\star}\right\| \\ &\leq (r+1)\left\|\vect\theta_{p} - \vect\theta_{\star}\right\|.
	\end{align}
	It follows that \eqref{eq:OverUpdateDecayCondition} holds if
	\begin{align}
	\alpha_V\left\|\vect\theta_{p+1} - \vect\theta_{p}\right\| + \zeta\left(\left\|\vect\theta_{p+1} - \vect\theta_{\star}\right\| - \left\|\vect\theta_{p} - \vect\theta_{\star}\right\|\right)& \leq \nonumber\\
	\leq  \alpha_V(r+1)\left\|\vect\theta_{p} - \vect\theta_{\star}\right\| + \zeta\left(r-1\right) \left\|\vect\theta_{p} - \vect\theta_{\star}\right\|& \leq  0.
	\end{align}
	Hence $W$ decreases when $\vect s_i$ lies outside of the level set $\mathbb{L}_{\vect{\theta}_{p+1}}$ if:
	\begin{align}
	\alpha_V (r+1) + \zeta\left(r-1\right) & \leq  0,
	\end{align}
	which can always be ensured by choosing:
	\begin{align}
	\zeta  \geq  \frac{\alpha_V  \left(r+1\right) }{1-r}.
	\end{align}
	As a result at all time $k$ whether a parameter update takes place or not, either function $W$ is decreasing or the system trajectory is contained in the level set $\mathbb{L}_{\vect\theta_p}$ corresponding to the MPC parameters in use. $\hfill\square$
\end{pf}

We elaborate in the next remarks on the assumptions and the stability claim made by Theorem \ref{Th:TrajectoryStability}.

\begin{Remark}  \label{Remark:WeakStability} 
	 The decrease of function $W$ at a time instant $i$ with corresponding parameter index $p$ is ensured for any $\vect s_i$ outside the level sets $\mathbb{L}_{\vect \theta_p}$, regardless of whether a parameter update has occurred or not. However, if $\vect s_i\in\mathbb{L}_{\vect \theta_p}$, then $W$ is not guaranteed to decrease, but the trajectory $\vect s_{i,i+1,\ldots}$ is guaranteed to remain in the level set $\mathbb{L}_{\vect \theta_p}$ until the next parameter update occurs.
	
Theorem \ref{Th:TrajectoryStability} guarantees the stabilization of the system trajectory in the state-parameter space, and under the Lyapunov function $W$, to the sequence of level sets $\mathbb{L}_{\vect \theta_p}$ converging to $\mathbb{L}_{\vect{\theta}_\infty}$. Theorem \ref{Th:TrajectoryStability} hence guarantees the stability of the system trajectory under fast parameter updates (e.g., at every sampling time $i$), albeit the parameter updates entering in the Lyapunov function via the norm $\Delta_p$ can temporarily drive the system trajectory away from the level sets (due to the fact that stability is guaranteed in the state-parameter update space, as opposed to the state space alone). 
	
	Hence a case covered by Theorem \ref{Th:TrajectoryStability} and expected in practice is one where the parameter updates are fairly slow and small compared to the system dynamics, possibly yielding a situation where the system trajectory is stabilized to $\mathbb{L}_{\vect{\theta}_\infty}$ mostly by moving from level set to level set, i.e., $\mathbb{L}_{\vect \theta_p}\rightarrow \mathbb{L}_{\vect \theta_{p+1}}\rightarrow \ldots\,$, without $W$ systematically decreasing. Theorem \ref{Th:TrajectoryStability}, however, guarantees that updating the parameters faster and more aggressively does not jeopardize the system stability.
	\end{Remark}

\begin{Remark}
	\label{rem:practical_stability}
	Let us further comment on Assumption~\ref{ass:convergence}, i.e., convergence of the parameters sequence $\vect\theta_{0,\ldots,\infty}$ delivered by the RL scheme. As previously discussed, many RL methods deliver a sequence of parameters that is stochastic by nature, because they are based on measurements taken from a stochastic system. We then observe that Equation~\eqref{eq:ThetaConvergence} is only satisfied asymptotically for large data sets. For RL methods based on very small data sets, such as, e.g., to the extreme those using basic stochastic gradient methods, one could consider an extension of Theorem \ref{Th:TrajectoryStability} where the decrease of $W$ holds only in a stochastic sense, or as practical stability or ISS. 
	To that end, one can, e.g., assume 
	\begin{align}
		\label{eq:ThetaConvergencePractical}
		\left\|\vect\theta_{p+1} - \vect\theta_{\star}\right\| \leq r\left\|\vect\theta_{p} - \vect\theta_{\star}\right\| + q,
	\end{align}
	which implies that the parameter updates converge to a neighborhood of zero. 
	The main conclusions of the Theorem still hold, \textit{mutatis mutandis}, with asymptotic stability replaced by practical stability. A formal discussion of this extension is not provided here for the sake of brevity, and we rather discuss next an alternative relaxation of Assumptions~\ref{ass:convergence}-\ref{ass:regularity}.
\end{Remark}

	While the result of Theorem~\ref{Th:TrajectoryStability} is strong, it also requires some assumptions that do not necessarily hold, and is typically only asymptotically valid. We provide next an alternative result which does not require Assumption~\ref{ass:convergence} and only requires a weaker version of Assumption~\ref{ass:regularity}, but also yields weaker conclusions.
	\begin{Assumption}
		\label{ass:Vregularity}
		It holds that 
		\begin{align}
			\label{eq:Vhat_regular}
			\hat V_{\vect\theta_{p+1}}\left(\vect s\right) - \hat V_{\vect\theta_p}\left(\vect s\right) \leq \beta(\Delta_p), 
		\end{align}
		for almost all $\vect{s}$ 
		with $\beta$ a $\mathcal{K}$ function.
	\end{Assumption}
Before stating the theorem, we observe that Assumption~\ref{ass:Vregularity} can be seen as a slight relaxation of Assumption~\ref{ass:regularity}.
\begin{theorem}
	\label{Th:iss}
	Suppose that Assumption~\ref{ass:Vregularity} holds and assume that each parameter is given by \eqref{eq:ConstRL:SQP} or \eqref{eq:ConstRL:FeasibleUpdates} such that $\vect\theta_{p}\in\vect\Theta_\mathrm{L} \cap \vect\Theta_\mathrm{F}\cap \vect\Theta_\mathcal{D}$. 
	 Then the closed-loop system is ISS.
\end{theorem}
\begin{pf}
	By assumption, we have that each $\hat V_{\vect\theta_p}$ is upper and lower bounded by $\mathcal{K}_\infty$ functions. Moreover,
	\begin{align*}
		\hat V_{\vect\theta_p}\left(\vect s_+\right) - \hat V_{\vect\theta_p}\left(\vect s\right) \leq -(1- \gamma) \hat V_{\vect\theta_p}\left(\vect s\right) + \delta_{\vect{\theta}_p},
	\end{align*}
	which, using~\eqref{eq:Vhat_regular}, entails that
	\begin{align*}
		\hat V_{\vect\theta_{p+1}}\left(\vect s_+\right) - \hat V_{\vect\theta_p}\left(\vect s\right) \leq -(1- \gamma) \hat V_{\vect\theta_p}\left(\vect s\right) + \delta_{\vect{\theta}_p} + \beta(\Delta_p),
	\end{align*}
	which is the decrease condition for ISS Lyapunov functions~\cite{Jiang2001}.
	Consequently, the closed-loop system is ISS with respect to $\Delta$ and $\delta_{\vect{\theta}}$. 	$\hfill\square$
\end{pf}
	
	The result of Theorem~\ref{Th:iss} can be understood as follows: while parameter updates might perturb the closed-loop system and temporarily jeopardize asymptotic stability, the destabilizing effect is bounded and disappears as soon as the parameters are not updated anymore. Consequently, as RL converges the possibly destabilizing perturbations decrease in intensity and eventually vanish. The original stability result of~\cite{Rawlings2017} is then recovered.

\section{Numerical Examples}
\label{sec:simulations}

In this section we provide numerical examples which illustrate the theoretical developments. 

\subsection{Recursive Feasibility}

We first discuss a simple academic example which is constructed, but allows us to discuss the theory in simple terms. 
Consider the scalar linear system
\begin{align*}
s_+ = As + Ba + w, && A = 1.1, && B = 0.1, 
\end{align*}
with $w\in[\underline{w},\overline{w}]:=[-0.1,0.1]$. 
We construct MPC such that it delivers a policy as close as possible to $-Ks+a^\mathrm{s}$, where $\vect \theta = \left\{K, a^\mathrm{s}\right\}$ are parameters to be adjusted by RL and the state and input must satisfy
\begin{align*}
s \leq \overline{s}:=0.1, && a\in [ -10, 10-0.5K ].
\end{align*}
One can verify that the robust MPC formulation
\begin{align*}
	\min_{u} \ \ & (u - (a^\mathrm{s}-Ks) )^2 \\
	\mathrm{s.t.} \ \ & As + Bu + \overline{w} \leq \overline{s}, \\
	& u\in [ -10, 10-0.5K ],
\end{align*}
guarantees that the state constraint $s \leq \overline{s}$ is never violated with the given dynamics and process noise. 
The stage cost $\ell(s,a) = (s-40)^2 + 10^{-4}a^2$ should be minimized by RL, and the MPC region of attraction at convergence must include the interval $[s_\mathrm{b}^0,s_\mathrm{b}^1]=[0,0.1]$. 

We consider a discount factor $\gamma=0.9$ and solve the problem by applying constrained policy gradient to the exact total expected cost $J$.
At each policy gradient iteration $p$ we solve the problem
	\begin{align*}
	\vect{\theta}_{p+1}:=\arg\min_{\vect{\theta}} \ & 0.5\|\vect{\theta}-\vect{\theta}_p\|_2^2 + \alpha\nabla_{\vect{\theta}} J^\top (\vect{\theta}-\vect{\theta}_p) \hspace{-10em}&&\hspace{10em}\\
	\mathrm{s.t.} \ & As^j_\mathrm{b} + Ba_\mathrm{b}^j + \overline{w} \leq \overline{s}, && j = 0,1,\\
	& a_\mathrm{b}^j := \left [-Ks_\mathrm{b}^j +a^\mathrm{s} \right ]_{-10}^{10-0.5K},  && j = 0,1,\\
	& A-BK \in [-1+\epsilon,1-\epsilon],
	\end{align*}
where $[\cdot]_a^b:=\max( a, \min( \cdot, b) )$, $\epsilon=10^{-6}$, $\alpha = 1$, and we computed $\nabla_{\vect{\theta}} J$ using the deterministic policy gradient theorem to compute the gradient in an actor-critic framework~\cite{Silver2014}.

\begin{figure}
	\begin{center}
		\includegraphics[width=0.9\linewidth]{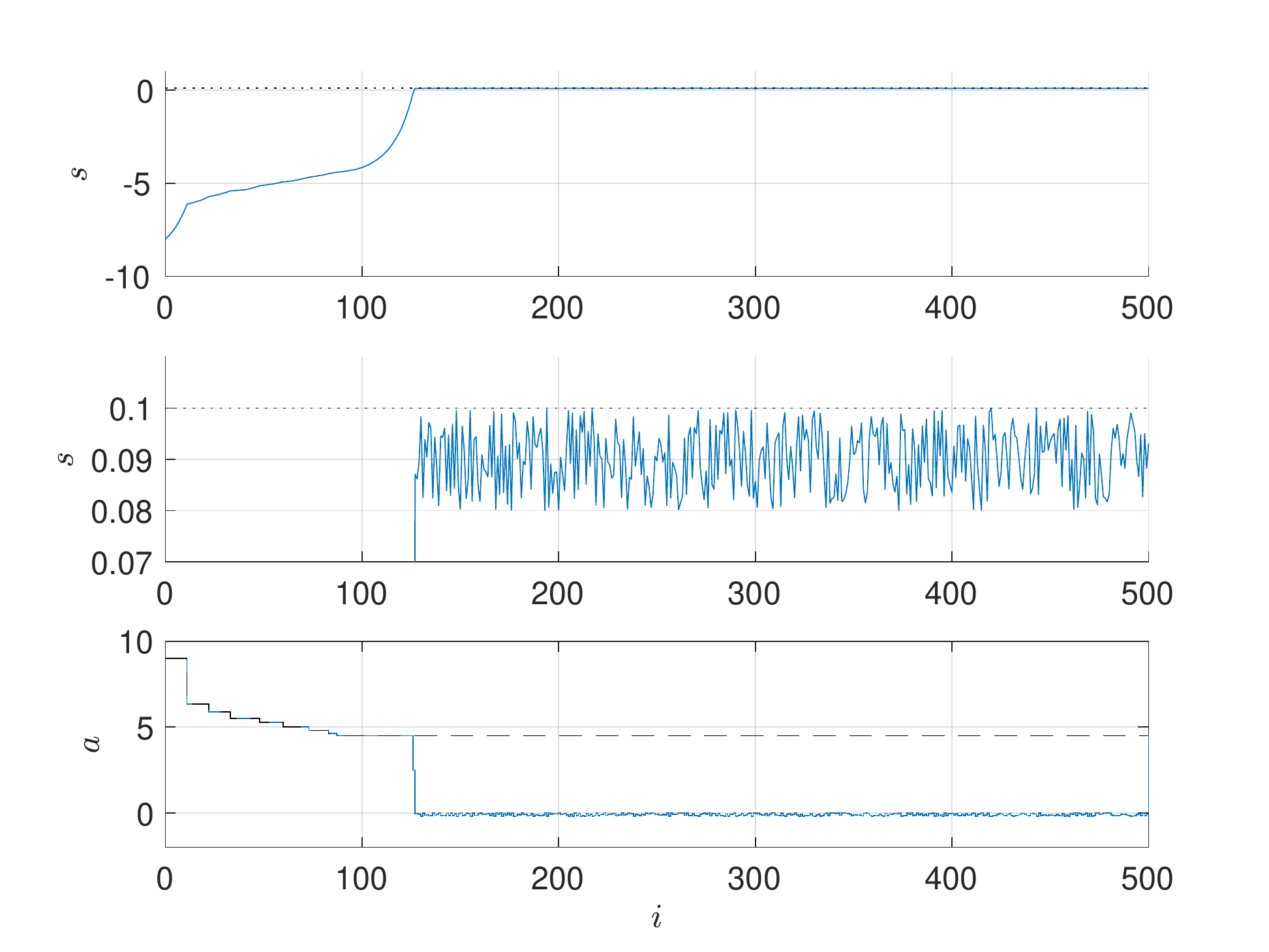}
		\caption{ 
		Closed-loop simulation with parameter updates relying on backtracking. Top plot: state trajectory. Middle plot: state trajectory zoom. Bottom plot: control trajectory (blue line) and upper bound $10-0.5K$ (dashed black line).}
		\label{fig:feasibility}
	\end{center}
\end{figure}

We initialize the problem with $K=2$, $a^\mathrm{s}=0$. The problem converges in one iterate to the optimal solution $K_\star = 11$, $a^\mathrm{s}=0.9$, but, depending on the initial state, the solution cannot be immediately applied to the system. Indeed, the region of attraction for the initial guess is the interval $\mathcal{S}_0:=[-8.9,0.1]$, while for the optimal solution the region of attraction is the interval $\mathcal{S}_\star:=[-4.4,0.1]$. We display in Figure~\ref{fig:feasibility} a simulation starting from $s=-8$ and using a backtracking strategy (Algorithm~\ref{Alg:SafeRLBacktrack}) with $n=1$, i.e., if the solution is not feasible, $\alpha$ is reduced with $\rho=0.9$. One can observe that, in the beginning, parameter $\vect{\theta}$ is not updated until: (a) $\alpha$ becomes smaller and, consequently, the region of attraction becomes larger; and (b) the state approaches the region of attraction. 

\begin{figure}
	\begin{center}
		\includegraphics[width=0.9\linewidth]{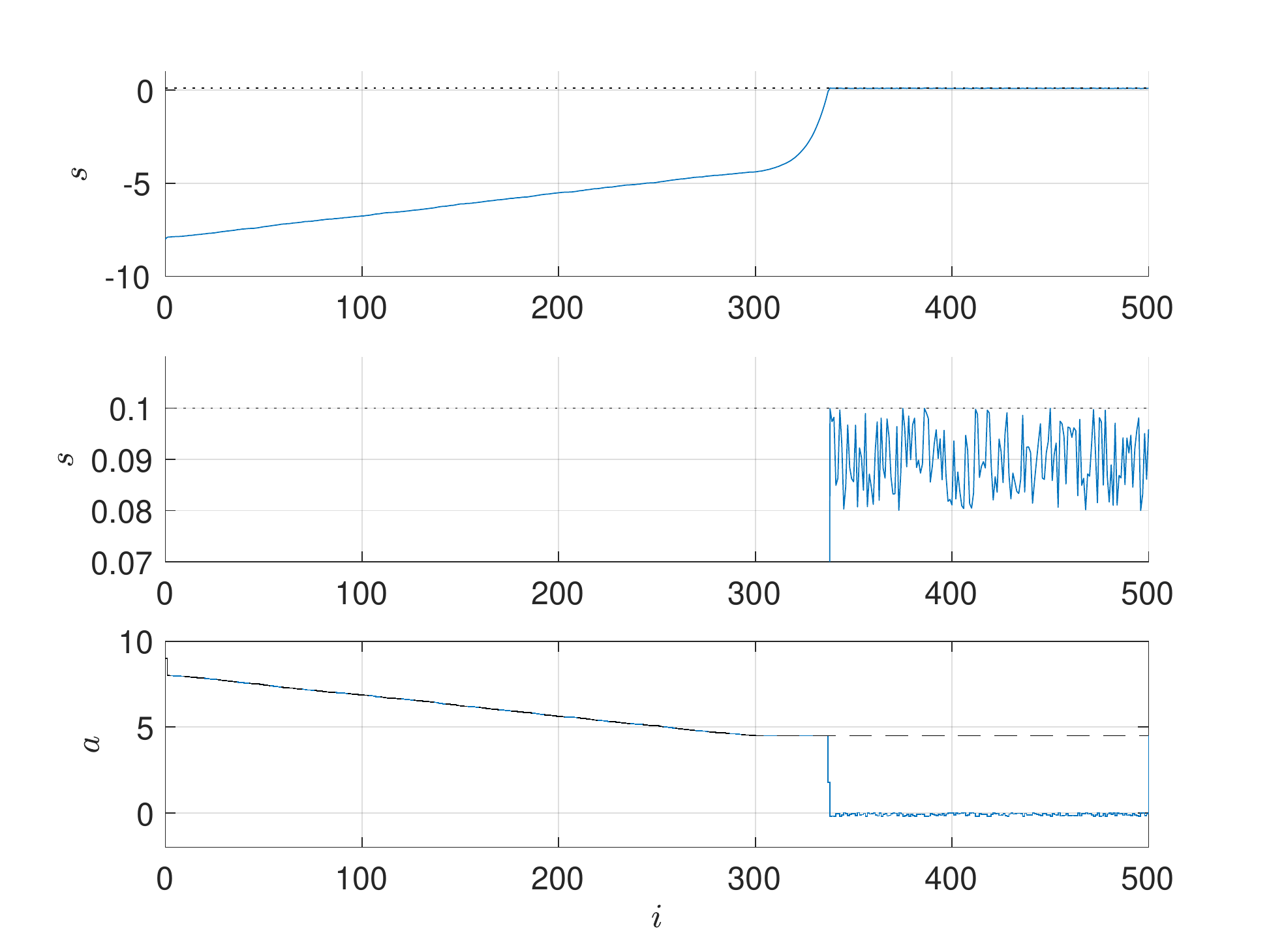}
		\caption{Closed-loop simulation with trajectory-constrained parameter updates. Top plot: state trajectory. Middle plot: state trajectory zoom. Bottom plot: control trajectory (blue line) and upper bound $10-0.5K$ (dashed black line).}
		\label{fig:feasibility2}
	\end{center}
\end{figure}

We also performed a simulation in which the update was done by trajectory-constrained parameter updates, where the following problem was solved
	\begin{align*}
	\vect{\theta}_{i+1}:=\arg\min_{\vect{\theta}} \ & 0.5\|\vect{\theta}-\vect{\theta}_i\|_2^2 + \alpha\nabla_{\vect{\theta}} J^\top (\vect{\theta}-\vect{\theta}_i) \hspace{-10em}&&\hspace{10em}\\
	\mathrm{s.t.} \ & As^j_\mathrm{b} + Ba_\mathrm{b}^j + \overline{w} \leq \overline{s}, && j = 0,1,\\
	& a_\mathrm{b}^j := \left [-K_i s_\mathrm{b}^j +a^\mathrm{s}_i \right ]_{-10}^{10-0.5K}, && j = 0,1,\\
	& A-BK \in [-1+\epsilon,1-\epsilon], \\
	& s_\mathrm{wc}^j \in \mathcal{S}_{\vect{\theta}}, && j = 0,1,
	\end{align*}
where $\mathcal{S}_{\vect{\theta}}$ is the region of attraction given $\vect{\theta}$, and $s_\mathrm{wc}^j$ are the one-step worst-case state realizations which, in this specific case, are given by
\begin{align*}
s_\mathrm{wc}^0 := As_i + B\left [-K_is_i +a_i^\mathrm{s} \right ]_{-10}^{10-0.5K_i} + \underline{w}, \\
s_\mathrm{wc}^1 := As_i + B\left [-K_is_i +a_i^\mathrm{s} \right ]_{-10}^{10-0.5K_i} + \overline{w}.
\end{align*}
The closed-loop trajectories are shown in Figure~\ref{fig:feasibility2}, where one can see that the convergence is slower than with backtracking, since the parameters are updated at each time, which reduces the maximum implementable control and, therefore, makes the convergence to the optimal operating set slower. Nevertheless, also in this case we recover the optimal solution $K_\infty = K_\star$, $a^\mathrm{s}_\infty=a^\mathrm{s}_\star$.

\subsection{Value Function}

We consider now the linear system with dynamics and stage cost
\begin{align*}
\vect{s}_+ &= \matr{cc}{1 & 0.1 \\ 0 & 1} \vect{s} + \matr{c}{0.05 \\ 0.1} \vect{a} + \vect{w}, \\
\ell(\vect{s},\vect{a}) &= \matr{c}{\vect{s}-\vect{s}^\mathrm{r} \\ \vect{a} -\vect{a}^\mathrm{r} }^\top \mathrm{diag}\left ( \matr{c}{1 \\ 0.01 \\ 0.01} \right ) \matr{c}{\vect{s}-\vect{s}^\mathrm{r} \\ \vect{a} -\vect{a}^\mathrm{r} },
\end{align*}
where $\vect{s}=(p,v)$ and $\vect{s}^\mathrm{r} =(-3,0)$, $\vect{a}^\mathrm{r}=0$.
We formulate a problem with prediction horizon $N=50$ and introduce the state and control constraints $-\vect{1}\leq \vect{s}\leq \vect{1}$, $-10\leq \vect{a}\leq 10$.
The real noise set is selected as a regular octagon, and we parametrize $\mathbb{W}_{\vect{\omega}}$ as a polytope with 4 facets.

We formulate tube based MPC as
\begin{subequations} %
	\label{eq:robust_mpc}%
	\begin{align}%
	Q_{\vect{\theta}}(\vect{s},\vect{a}):= \hspace{-1em}& \nonumber\\
	\min_{\vect{z}} \ \ & \sum_{k=0}^{N-1} \norm{\vect{x}_k - \vect{x}_\mathrm{r} \\ \vect{u}_k  - \vect{u}_\mathrm{r}  }^2_H 
	+ \norm{\vect{x}_N - \vect{x}_\mathrm{r}}^2_P 
	\nonumber \\
	&\hspace{6.5em}
	+ \norm{\vect{x}_0}^2_\Lambda + \vect{\lambda}^\top \vect{x}_0  + l
	\label{eq:robust_mpc_cost} \hspace{-10em}&\hspace{0em}\\ 
	\mathrm{s.t.} \ \ & \vect{x}_0 = \vect{s}, \qquad \vect{u}_0 = \vect{a}, \label{eq:robust_mpc_ic}\\
	& \vect{x}_{k+1} = A\vect{x}_k + B \vect{u}_k + \vect{b}, & \hspace{-3em} k\in\mathbb{I}_0^{N-1}, \label{eq:robust_mpc_dyn}\\
	& C\vect{x}_k + D \vect{u}_k + \vect{c}_k \leq \vect{0}, & \hspace{-3em} k\in\mathbb{I}_0^{N-1}, \label{eq:robust_mpc_pc}\\
	& G\vect{x}_N  + \vect{g} \leq \vect{0},\label{eq:robust_mpc_tc}
	\end{align}%
\end{subequations}%
where one must enforce that the system dynamics~\eqref{eq:robust_mpc_dyn} and a parametrized compact uncertainty set $\mathbb{W}_{\vect{\omega}}$ are such that
$
\vect{s}_+ - (A\vect{s}+B\vect{a} + \vect{b}) \in \mathbb{W}_{\vect{\omega}}.
$
This issue has been discussed in detail in~\cite{Zanon2021}, where the set is parametrized as the polyhedron $\mathbb{W}_{\vect{\omega}}:=\{ \ \vect{w} \ | \ M\vect{w} \leq \vect{m} \ \}$ and the following set membership constraint is imposed on $\vect\omega=(M,\vect{m})$ for all past samples $\vect{s}_{i+1}, \vect{s}_i, \vect{a}_i$, $i\in\mathcal{I}$:
\begin{align*}
M(\vect{s}_{i+1} - (A\vect{s}_i+B\vect{a}_i + \vect{b})) \leq \vect{m}, && \forall \ i\in\mathcal{I}.
\end{align*}
Then, $\vect{c}_k$ is computed by tightening the original constraints
$
C\vect{s} + D\vect{a} + \vect{\hat c} \leq 0,
$
so as to guarantee that, for any process noise $\vect{w} \in \mathbb{W}_{\vect{\omega}}$, the constraints are satisfied. 
Moreover, parameters $\vect{x}_\mathrm{r},\vect{u}_\mathrm{r}$ must be a steady-state for the system dynamics~\eqref{eq:robust_mpc_dyn}, i.e., 
\begin{align*}
(A-I) \vect{x}_\mathrm{r} + B \vect{u}_\mathrm{r} = \vect{0}.
\end{align*}
Finally, $G$ and $\vect{g}$ must be selected such that they define a robust positively invariant terminal set for the feedback law $\vect{u}=-K(\vect{x}-\vect{x}_\mathrm{r}) + \vect{u}_\mathrm{r}$, with $K$ the solution to the LQR formulated with $A,B,H,P$. 
The vector of MPC parameters is then defined as
\begin{align}
\vect{\theta} = \{ \Lambda, \lambda, l, H, \vect{x}_\mathrm{r}, \vect{u}_\mathrm{r}, M \},
\end{align}
and we consider $K$, $P$, $\vect{c}_k$, $G$, $\vect{g}$ as functions of these parameters. Vector $\vect{m}$ can also be included in $\vect{\theta}$, but, as discussed in~\cite{Zanon2021} this is not necessary. Matrices $C$, $D$ and vector $\vect{\bar{c}}$ are assumed to be known. Finally, $A$, $B$, $\vect{b}$ could in principle also be included in the parameter vector $\vect{\theta}$. However, as discussed in~\cite{Zanon2021} this makes the safe RL problem much harder to formulate and solve, since it obliges one to store very large amounts of data and formulate an equally large amount of constraints. 

The set of parameters guaranteeing safety and stability then becomes
\begin{align*}
\Theta := \{ \ \vect{\theta} \ | \ & H\succ 0, \\
& M(\vect{s}_{i+1} - (A\vect{s}_i+B\vect{a}_i + \vect{b})) \leq \vect{m}, \ \forall \ i \in \mathcal{I}, \\
& (A-I) \vect{x}_\mathrm{r} + B \vect{u}_\mathrm{r} = \vect{0}, \\
&\exists \ \vect{x} \ \mathrm{s.t.} \ G\vect{x} \leq \vect{g}
\ \},
\end{align*}
i.e., the noise set must include all observed noise samples, the reference must be a steady-state of the system and the terminal set must be nonempty. This last condition also entails that the MPC domain is nonempty.

We update $\vect{\theta}$ using a batch $Q$ learning approach with batches of horizon $N_\mathrm{b}=20$ with learning rate $\alpha=0.1$, using the backtracking strategy.

\begin{figure}
	\includegraphics[width=\linewidth]{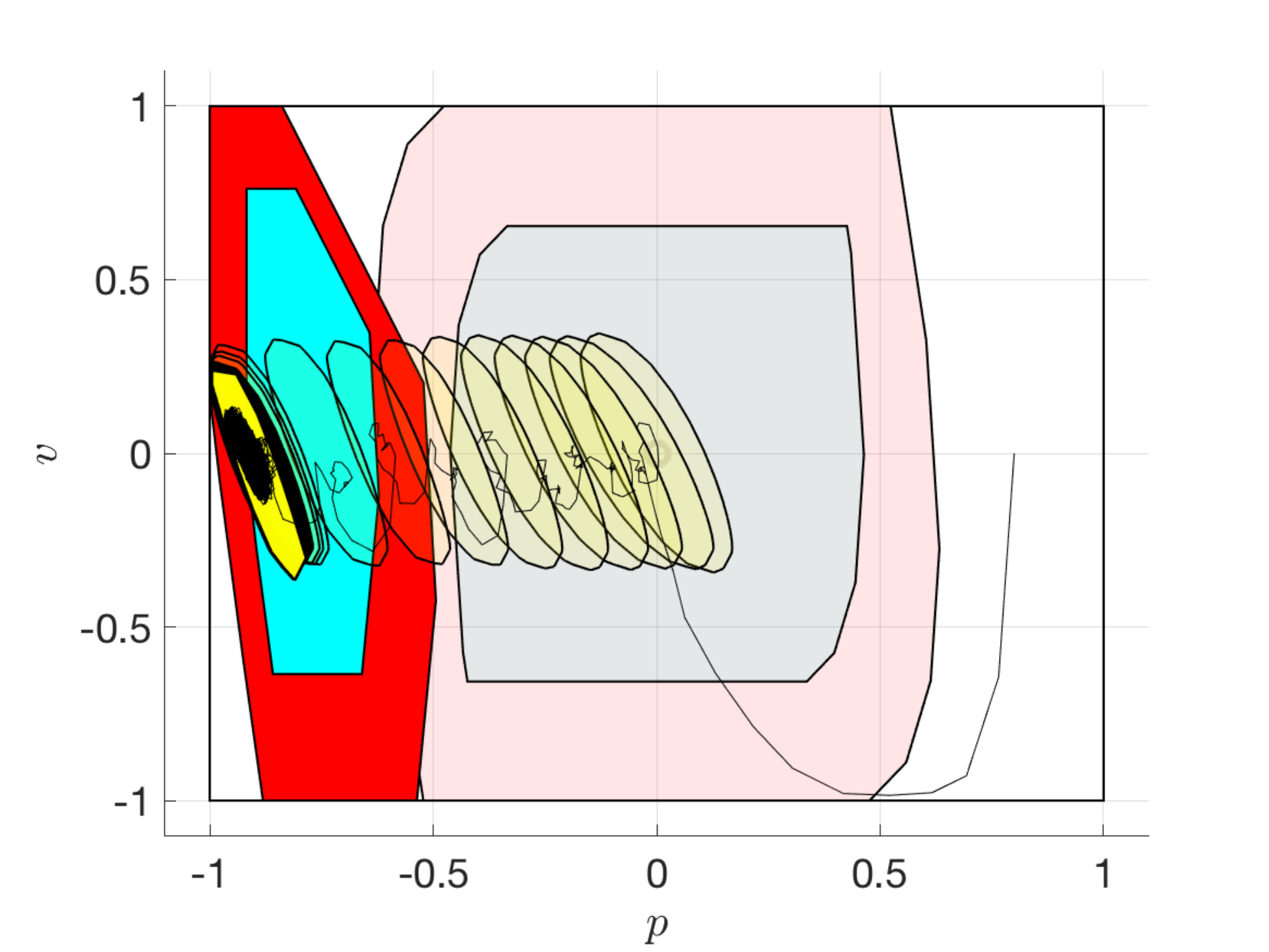}
	\caption{MRPI (red), terminal (cyan) sets and reference $\vect{x}^\mathrm{r}$ (black and grey circle) at the beginning and end of the learning process; state trajectoyr (black line) and mRPI sets (yellow) at each time instant.}
	\label{fig:tube_sets}
\end{figure}

\begin{figure}
	\includegraphics[width=\linewidth]{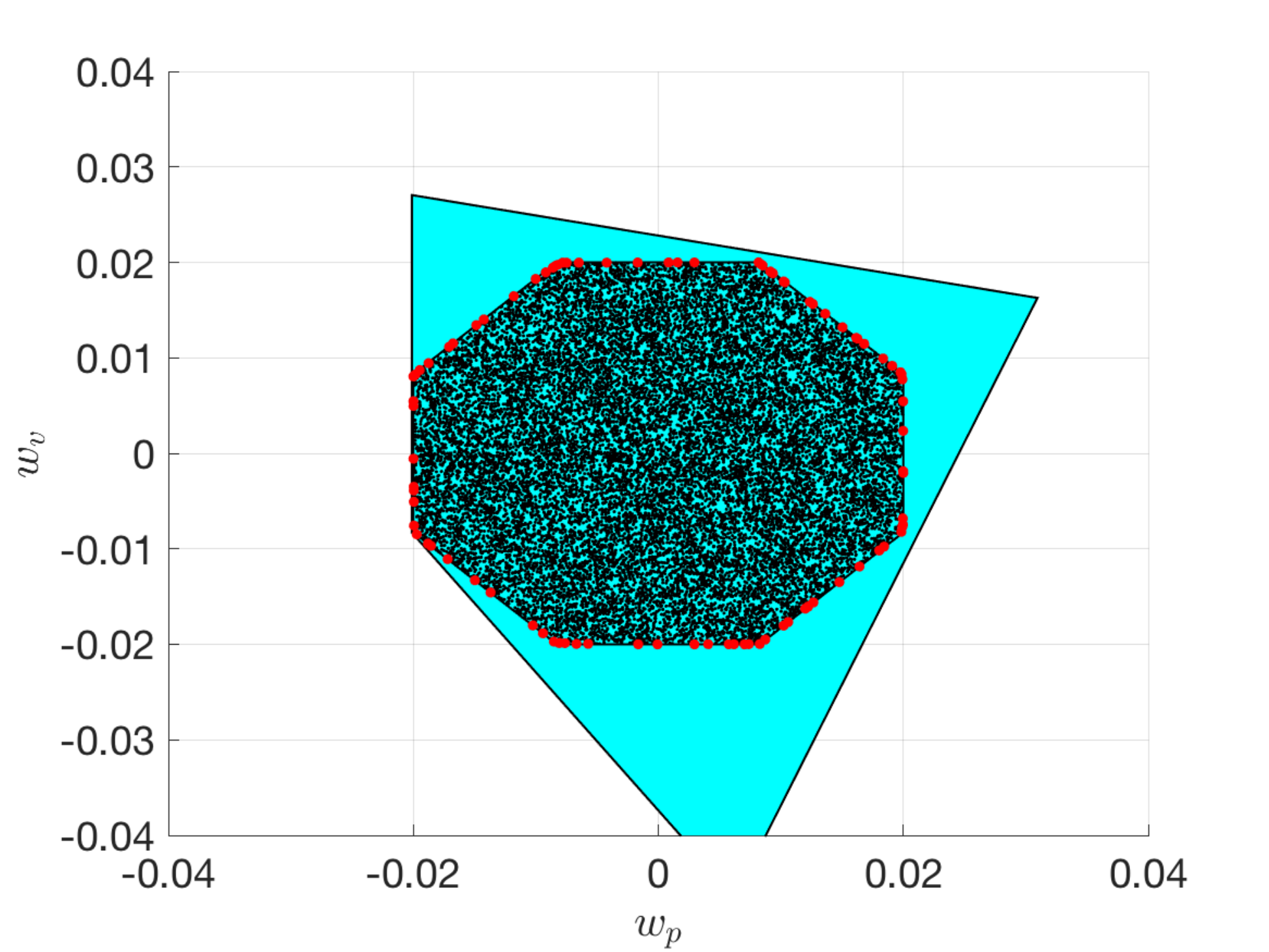}
	\caption{True process noise set (transparent octogon), noise samples (black dots), their convex hull (red dots) and noise set parametrized by matrix $M$ (cyan).}
	\label{fig:tube_noise}
\end{figure}

\begin{figure}
	\includegraphics[width=\linewidth]{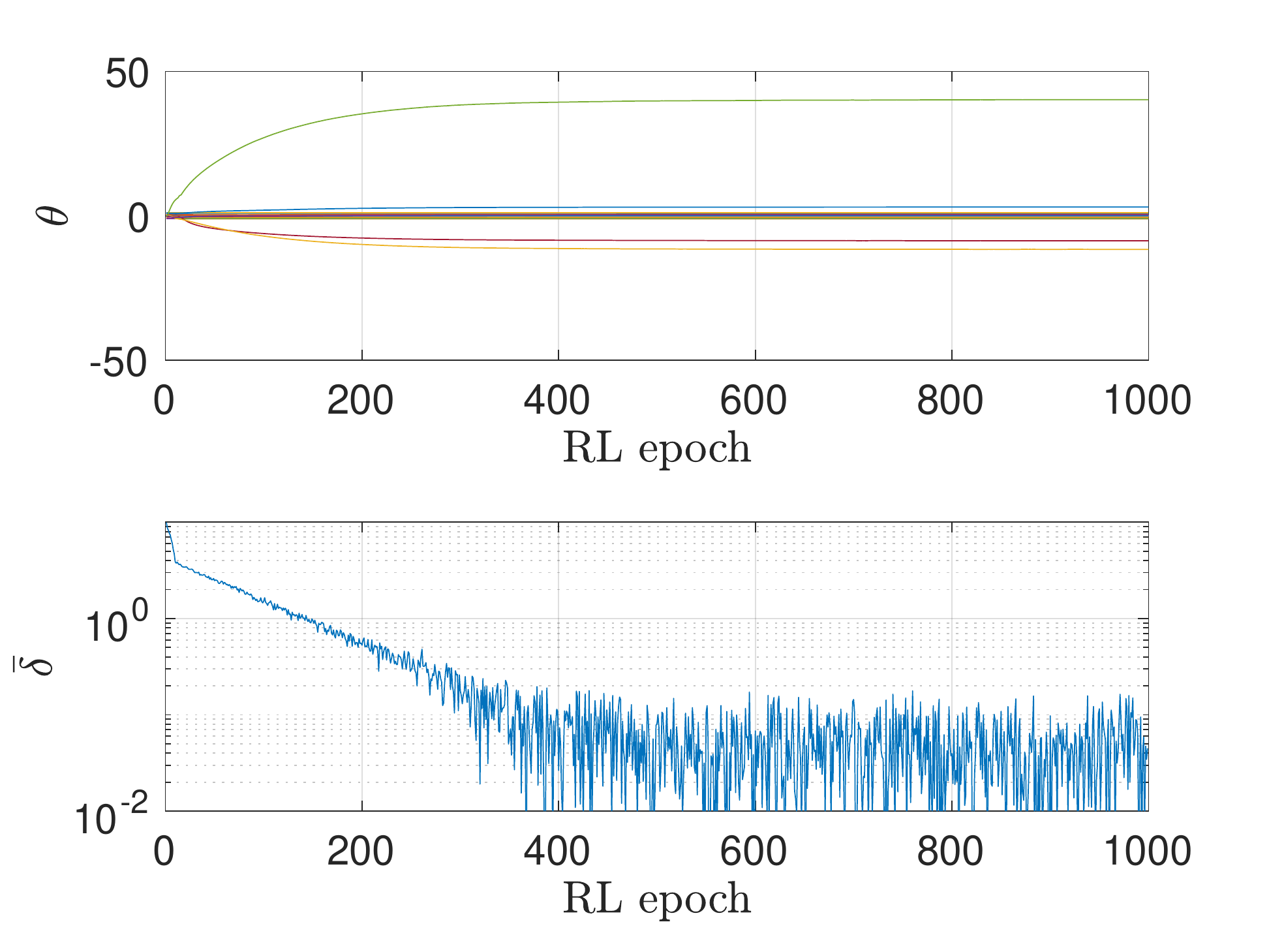}
	\caption{Top plot: parameter evolution through the epochs. Bottom plot: TD error through the epochs.}
	\label{fig:tube_params}
\end{figure}

\begin{figure}
	\includegraphics[width=\linewidth]{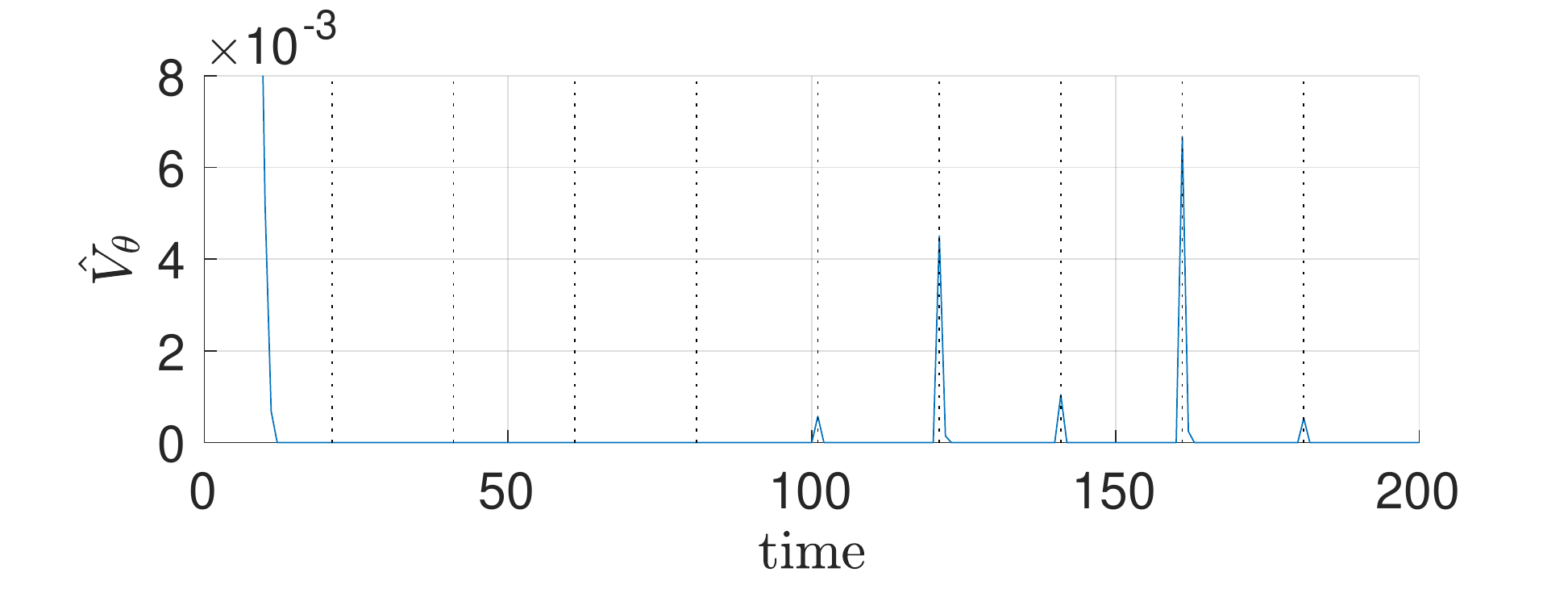}
	\includegraphics[width=\linewidth]{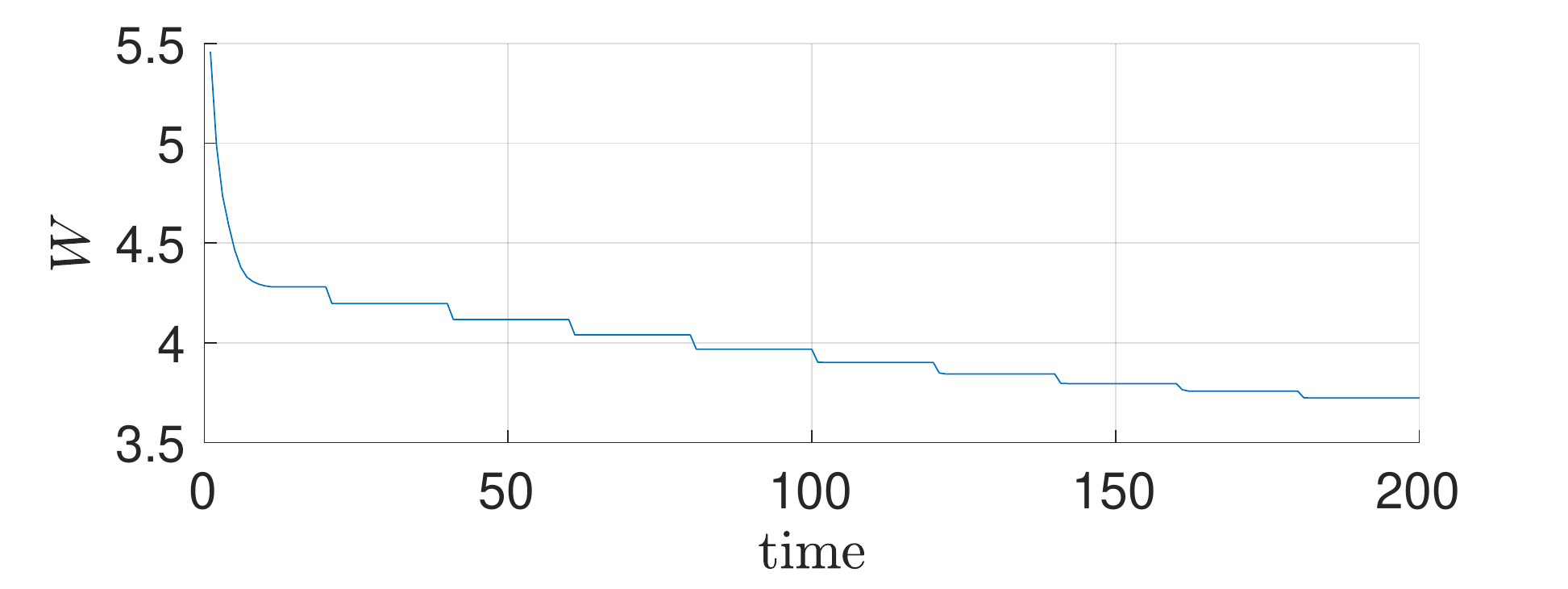}
	\caption{Top figure: Lyapunov function $\hat V_{\vect{\theta}}$ over the first epochs. Bottom plot: Lyapunov function $W$ over time.}
	\label{fig:tube_lyap}
\end{figure}

\begin{figure}
	\includegraphics[width=\linewidth]{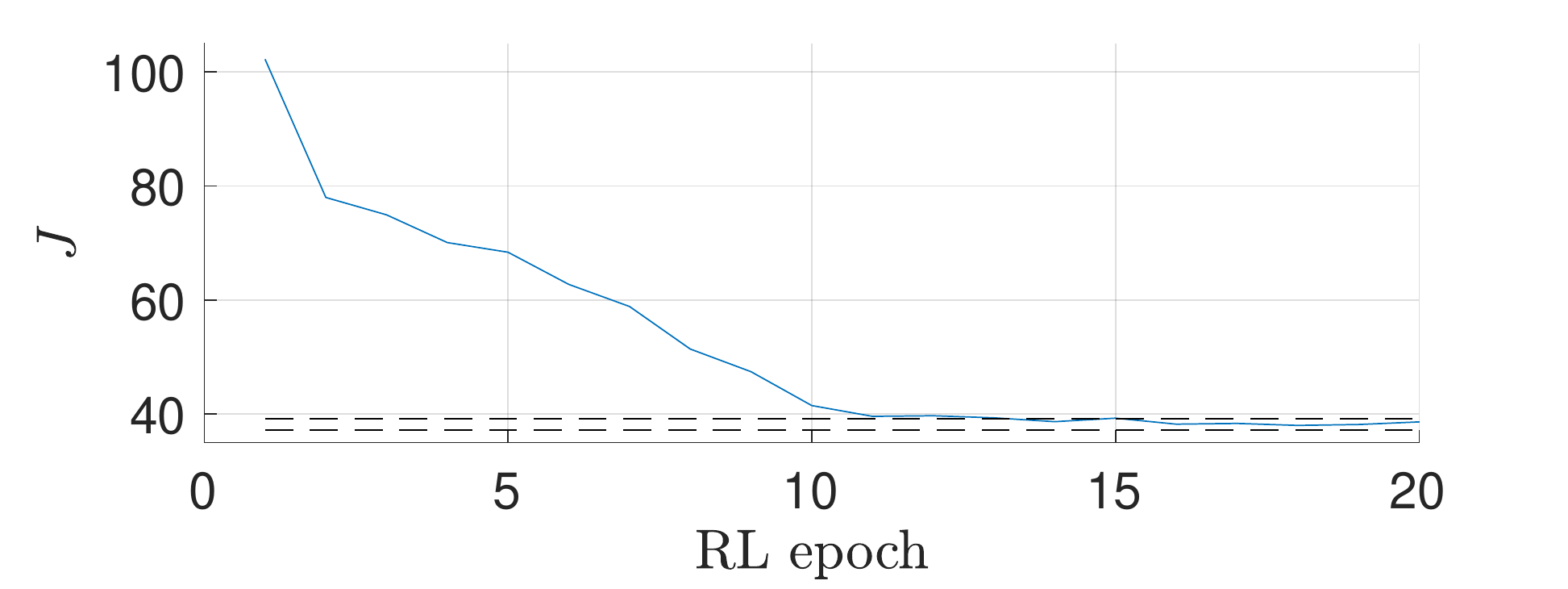}
	\caption{Performance $J$ in terms of total discounted cost for each batch. The dashed lines indicate the maximum and minimum of $J$ over all future times.}
	\label{fig:tube_performance}
\end{figure}

We  simulated the system starting from state $\vect{s}_0=(0.8,0)$. The backtracking strategy never rejected nor reduced any step. The resulting closed-loop trajectory is displayed in Figure~\ref{fig:tube_sets}, together with the reference, maximum robust positive invariant (MRPI) and terminal sets at the beginning and end of the simulation, as well as the minimum robust positive invariant (mRPI) sets throughtout the simulation. We display the noise set approximation at the end of the simulation in Figure~\ref{fig:tube_noise}, and the evolution throughout the RL epochs of the parameter $\vect{\theta}$ and the average TD error in each batch in Figure~\ref{fig:tube_params}. We display the MPC Lyapunov functions $\hat V_{\vect{\theta}}$ and $W$ in time in Figure~\ref{fig:tube_lyap}. One can see that in the beginning $\hat V_{\vect{\theta}}$ sometimes increases upon parameter updates, but decreases inside each batch. Note that this result is perfectly in line with Theorem~\ref{Th:TrajectoryStability} and Remark~\ref{Remark:WeakStability}. After the displayed time interval, the Lyapunov function $\hat V_{\vect{\theta}}$ was always 0, i.e., the state trajectory remained inside the mRPI set, even when this set was updated by a parameter change. Some words are due in order to discuss function $W$: as pointed out in Remark~\ref{rem:practical_stability}, in practice one can at best expect that the parameters converge to a neighborhood of the optimal ones. Therefore, we selected $\vect{\theta}_\star$ as the average of $\vect{\theta}$ over the last $100$ epochs, when, as shown in Figure~\ref{fig:tube_params}, the parameter is at convergence. We observed that $\zeta=0.1$ was sufficiently high to satisfy the conditions of Theorem~\ref{Th:TrajectoryStability}. As one can see in Figure~\ref{fig:tube_lyap}, differently from $\hat V_{\vect{\theta}}$, the obtained $W$ is decreasing also in the first epochs. 
 Finally, we show the performance $J$ in terms of total discounted cost over each batch in Figure~\ref{fig:tube_performance}. One can see that after a short transient the performance reaches convergence and does not improve anymore.

\section{Conclusions}
\label{sec:conclusions}
This paper discusses how to implement Learning-based adaptations of a robust MPC scheme in order to improve its closed-loop performance while maintaining the stability and safety of the control policy. We show in particular that these requirements can be treated via constraints on the learning steps, and parameter update conditions that are fairly simple to verify, and that can be implemented online in real time. We additionally establish that the proposed approach ensures that the update conditions are not blocking the learning process, in the sense that they are met in final time with probability one. We finally show that under some conditions on the learning process, a form of stability of the resulting learning-based robust MPC scheme is guaranteed in the state-parameter space. The proposed approaches are illustrated in two simulated examples.

\bibliographystyle{plain}
\bibliography{syscop}

\begin{thebibliography}{10}

\bibitem{Abdufattokhov2021}
S.~Abdufattokhov, M.~Zanon, and A.~Bemporad.
\newblock {L}earning {C}onvex {T}erminal {C}osts for {C}omplexity {R}eduction
  in {MPC}.
\newblock In {\em 2021 60th IEEE Conference on Decision and Control (CDC)},
  pages 2163--2168. 2021.

\bibitem{Amos2018}
B.~Amos, I.~D.~J. Rodriguez, J.~Sacks, B.~Boots, and J.~Z. Kolter.
\newblock Differentiable mpc for end-to-end planning and control.
\newblock In {\em Proceedings of NIPS}, NIPS'18, pages 8299--8310, USA, 2018.
  Curran Associates Inc.

\bibitem{Aswani2013}
A.~Aswani, H.o Gonzalez, S.~S. Sastry, and C.~Tomlin.
\newblock {P}rovably safe and robust learning-based model predictive control.
\newblock {\em Automatica}, 49(5):1216 -- 1226, 2013.

\bibitem{Berkenkamp2017}
F.~Berkenkamp, M.~Turchetta, A.~Schoellig, and A.~Krause.
\newblock {S}afe {M}odel-based {R}einforcement {L}earning with {S}tability
  {G}uarantees.
\newblock In I.~Guyon, U.~V. Luxburg, S.~Bengio, H.~Wallach, R.~Fergus,
  S.~Vishwanathan, and R.~Garnett, editors, {\em Advances in Neural Information
  Processing Systems 30}, pages 908--918. Curran Associates, Inc., 2017.

\bibitem{Bertsekas1971a}
D.P. Bertsekas and I.B. Rhodes.
\newblock {Recursive state estimation for a set-membership description of
  uncertainty}.
\newblock {\em IEEE Transactions on Automatic Control}, 16:117--128, 1971.

\bibitem{Chisci2001}
L.~Chisci, J.A. Rossiter, and G.~Zappa.
\newblock {S}ystems with persistent disturbances: predictive control with
  restricted constraints.
\newblock {\em Automatica}, 37:1019--1028, 2001.

\bibitem{Fiacco1983}
A.V. Fiacco.
\newblock {\em {I}ntroduction to sensitivity and stability analysis in
  nonlinear programming}.
\newblock Academic Press, New York, 1983.

\bibitem{Gaskett2003}
Chris Gaskett.
\newblock {R}einforcement learning under circumstances beyond its control.
\newblock In {\em International conference on computational intelligence,
  robotics and autonomous systems}, 2003.

\bibitem{Gros2020}
S.~{Gros} and M.~{Zanon}.
\newblock {D}ata-{D}riven {E}conomic {NMPC} {U}sing {R}einforcement {L}earning.
\newblock {\em IEEE Transactions on Automatic Control}, 65(2):636--648, Feb
  2020.

\bibitem{Heger1994}
Matthias Heger.
\newblock {C}onsideration of {R}isk in {R}einforcement {L}earning.
\newblock In {\em Machine Learning Proceedings 1994}, pages 105--111. Elsevier,
  1994.

\bibitem{Hewing2020}
Lukas Hewing, Kim~P. Wabersich, Marcel Menner, and Melanie~N. Zeilinger.
\newblock {L}earning-{B}ased {M}odel {P}redictive {C}ontrol: {T}oward {S}afe
  {L}earning in {C}ontrol.
\newblock {\em Annual Review of Control, Robotics, and Autonomous Systems},
  3(1):269--296, 2020.

\bibitem{Jiang2001}
Zhong-Ping Jiang and Yuan Wang.
\newblock Input-to-state stability for discrete-time nonlinear systems.
\newblock {\em Automatica}, 37(6):857--869, 2001.

\bibitem{Koehler2021}
Johannes K\"ohler, Peter K\"otting, Raffaele Soloperto, Frank Allg\"ower, and
  Matthias~A. M\"uller.
\newblock {A} {R}obust {A}daptive {M}odel {P}redictive {C}ontrol {F}ramework
  for {N}onlinear {U}ncertain {S}ystems.
\newblock {\em International Journal of Robust and Nonlinear Control},
  31(18):8725--8749, 2021.

\bibitem{Koller2018}
T.~Koller, F.~Berkenkamp, M.~Turchetta, and A.~Krause.
\newblock {L}earning-based {M}odel {P}redictive {C}ontrol for {S}afe
  {E}xploration and {R}einforcement {L}earning.
\newblock Published on Arxiv, 2018.

\bibitem{Lewis2009}
F.~L. Lewis and D.~Vrabie.
\newblock Reinforcement learning and adaptive dynamic programming for feedback
  control.
\newblock {\em IEEE Circuits and Systems Magazine}, 9(3):32--50, 2009.

\bibitem{Lewis2012}
F.~L. Lewis, D.~Vrabie, and K.~G. Vamvoudakis.
\newblock Reinforcement learning and feedback control: Using natural decision
  methods to design optimal adaptive controllers.
\newblock {\em IEEE Control Systems}, 32(6):76--105, 2012.

\bibitem{Limon2008}
D.~Limon, I.~Alvarado, T.~Alamo, and E.F. Camacho.
\newblock {MPC} for {T}racking {P}iecewise {C}onstant {R}eferences for
  {C}onstrained {L}inear {S}ystems.
\newblock {\em Automatica}, 44(9):2382--2387, 2008.

\bibitem{Masti2022}
D.~Masti, M.~Zanon, and A.~Bemporad.
\newblock {T}uning {LQR} {C}ontrollers: a {S}ensitivity-{B}ased {A}pproach.
\newblock {\em IEEE Control Systems Letters}, 6:932--937, 2022.
\newblock Also in 60th IEEE Conf. on Decision and Control, Austin, TX, December
  13-15, 2021.

\bibitem{Mayne2011}
D.~Q. Mayne, E.~C. Kerrigan, E.~J. van Wyk, and P.~Falugi.
\newblock Tube-based robust nonlinear model predictive control.
\newblock {\em International Journal of Robust and Nonlinear Control},
  21(11):1341--1353, 2011.

\bibitem{Mayne2005}
D.Q. Mayne, M.M. Seron, and S.V. Rakovic.
\newblock {R}obust model predictive control of constrained linear systems with
  bounded disturbances.
\newblock {\em Automatica}, 41:219--224, 2005.

\bibitem{Murray2018}
R.~Murray and M.~Palladino.
\newblock A model for system uncertainty in reinforcement learning.
\newblock {\em Systems \& Control Letters}, 122:24 -- 31, 2018.

\bibitem{Nocedal2006}
J.~Nocedal and S.J. Wright.
\newblock {\em {N}umerical {O}ptimization}.
\newblock Springer Series in Operations Research and Financial Engineering.
  Springer, 2 edition, 2006.

\bibitem{Ostafew2016}
C.~J. Ostafew, A.~P. Schoellig, and T.~D. Barfoot.
\newblock {R}obust {C}onstrained {L}earning-based {NMPC} enabling reliable
  mobile robot path tracking.
\newblock {\em The International Journal of Robotics Research},
  35(13):1547--1563, 2016.

\bibitem{Papaioannou2015}
Iason Papaioannou, Wolfgang Betz, Kilian Zwirglmaier, and Daniel Straub.
\newblock {MCMC} algorithms for {S}ubset {S}imulation.
\newblock {\em Probabilistic Engineering Mechanics}, 41:89--103, 2015.

\bibitem{Rawlings2017}
J.~B. Rawlings, D.~Q. Mayne, and M.~Diehl.
\newblock {\em {M}odel {P}redictive {C}ontrol: {T}heory, {C}omputation, and
  {D}esign}.
\newblock Nob Hill Publishing, 2nd edition, 2017.

\bibitem{Silver2014}
D.~Silver, G.~Lever, N.~Heess, T.~Degris, D.~Wierstra, and M.~Riedmiller.
\newblock Deterministic policy gradient algorithms.
\newblock In {\em Proceedings of ICML}, ICML'14, pages I--387--I--395, 2014.

\bibitem{Soloperto2023}
R.~Soloperto, M.~M\"uller, and F.~Allg\"ower.
\newblock {G}uaranteed {C}losed-{L}oop {L}earning in {M}odel {P}redictive
  {C}ontrol.

\bibitem{Sutton1999}
R.~S. Sutton, D.~McAllester, S.~Singh, and Y.~Mansour.
\newblock Policy gradient methods for reinforcement learning with function
  approximation.
\newblock In {\em Proceedings of NIPS}, pages 1057--1063, Cambridge, MA, USA,
  1999. MIT Press.

\bibitem{Villanueva2017}
Mario~E. Villanueva, Rien Quirynen, Moritz Diehl, Beno\^it Chachuat, and Boris
  Houska.
\newblock {R}obust {MPC} via {M}in-{M}ax {D}ifferential {I}nequalities.
\newblock {\em Automatica}, 77:311--321, 2017.

\bibitem{Wabersich2019}
K.~Wabersich, L.~Hewing, A.~Carron, and M.~Zeilinger.
\newblock Probabilistic model predictive safety certification for
  learning-based control.
\newblock {\em arXiv:1906.10417v1, 25 Jun 2019}, 2019.

\bibitem{Zanon2021}
M.~Zanon and Gros.
\newblock {S}afe {R}einforcement {L}earning {U}sing {R}obust {MPC}.
\newblock {\em Transaction on Automatic Control}, 66(8):3638--3652, 2021.

\bibitem{Zanon2021a}
M.~Zanon and S.~Gros.
\newblock {O}n the {S}imilarity {B}etween {T}wo {P}opular {T}ube {MPC}
  {F}ormulations.
\newblock In {\em Proceedings of the European Control Conference}, pages
  651--656, 2021.

\bibitem{Zanon2019}
M.~Zanon, S.~Gros, and A.~Bemporad.
\newblock {P}ractical {R}einforcement {L}earning of {S}tabilizing {E}conomic
  {MPC}.
\newblock In {\em Proceedings of the European Control Conference}, pages
  2258--2263, 2019.

\bibitem{Zhu2022}
Mengjia Zhu, Dario Piga, and Alberto Bemporad.
\newblock {C-GLISp}: {P}reference-{B}ased {G}lobal {O}ptimization {U}nder
  {U}nknown {C}onstraints {W}ith {A}pplications to {C}ontroller {C}alibration.
\newblock {\em IEEE Transactions on Control Systems Technology}, 2022.
\newblock (in press).

\bibitem{Zuev2021}
Konstantin~M. Zuev.
\newblock {\em {S}ubset {S}imulation {M}ethod for {R}are {E}vent {E}stimation:
  {A}n {I}ntroduction}, pages 1--25.
\newblock Springer Berlin Heidelberg, Berlin, Heidelberg, 2021.

\end{thebibliography}

\end{document}